\documentclass[journal]{IEEEtran}

\ifCLASSINFOpdf
\else
\fi

\usepackage{times}
\usepackage{epsfig}
\usepackage{graphicx}
\usepackage{amsmath}
\usepackage{amssymb}
\usepackage{graphicx}
\usepackage{cite}
\usepackage{enumerate}
\usepackage{cases}
\usepackage{multirow}
\usepackage{verbatim}
\usepackage{amssymb}
\usepackage{CJK}
\usepackage{algorithm}
\usepackage{algorithmicx}
\usepackage{algpseudocode}

\usepackage{color}
\usepackage{bm}
\usepackage{booktabs}
\usepackage[colorlinks,linkcolor=red]{hyperref}

\newcommand{\ie}{\textit{i}.\textit{e}.}
\newcommand{\eg}{\textit{e}.\textit{g}.}
\newcommand{\etc}{\textit{etc}.}

\begin{document}

\title{Nested Network with Two-Stream Pyramid\\ for Salient Object Detection in Optical\\ Remote Sensing Images}


\author{Chongyi Li, Runmin Cong, Junhui Hou,~\IEEEmembership{Member,~IEEE,} Sanyi Zhang, Yue Qian, and Sam Kwong,~\IEEEmembership{Fellow,~IEEE}
\thanks{Manuscript received Nov. 2018. This work was supported in part by the National Natural Science Foundation of China under Grant 61871342, in part by Hong Kong RGC General Research Funds under Grant 9042038 (CityU 11205314) and Grant 9042322 (CityU 11200116), and in part by Hong Kong RGC Early Career Schemes under Grant 9048123 (CityU 21211518). (\emph{Chongyi Li and Runmin Cong contributed equally to this work. Corresponding author: Runmin Cong})}
\thanks{C. Li and Y. Qian are with the Department of Computer Science, City University of Hong Kong, Kowloon 999077, Hong Kong  (e-mail: lichongyi25@gmail.com, yueqian4-c@my.cityu.edu.hk).}
\thanks{R. Cong is the Institute of Information Science, Beijing Jiaotong University, Beijing 100044, China, and also with the Beijing Key Laboratory of Advanced Information Science and Network Technology, Beijing Jiaotong University, Beijing 100044, China (e-mail: rmcong@126.com).}
\thanks{J. Hou and S. Kwong are with the Department of Computer Science, City University of Hong Kong, Kowloon 999077, Hong Kong, and also with the City University of Hong Kong Shenzhen Research Institute, Shenzhen 51800, China (e-mail: jh.hou@cityu.edu.hk, cssamk@cityu.edu.hk).}
\thanks{S. Zhang is with the School of Electrical and Information Engineering, Tianjin University, Tianjin 300072, China (e-mail: zhangsanyi@tju.edu.cn).}

}

\markboth{IEEE TRANSACTIONS ON GEOSCIENCE AND REMOTE SENSING}%
{Shell \MakeLowercase{\textit{et al.}}: Bare Demo of IEEEtran.cls for IEEE Journals}

\maketitle

\begin{abstract}
Arising from the various object types and scales, diverse imaging orientations, and cluttered backgrounds in optical remote sensing image (RSI), it is difficult to directly extend the success of salient object detection for nature scene image to the optical RSI. In this paper, we propose an end-to-end deep network called \textbf{LV-Net} based on the shape of network architecture, which detects salient objects from optical RSIs in a purely data-driven fashion. The proposed LV-Net consists of two key modules, \ie, a two-stream pyramid module (L-shaped module) and an encoder-decoder module with nested connections (V-shaped module). Specifically, the L-shaped module extracts a set of complementary information  hierarchically by using a two-stream pyramid structure, which is beneficial to perceiving the diverse scales and local details of salient objects. The V-shaped module gradually integrates encoder detail features with decoder semantic features through nested connections, which aims at suppressing the cluttered backgrounds and highlighting the salient objects. In addition, we construct the first publicly available optical RSI dataset for salient object detection, including 800 images with varying spatial resolutions, diverse saliency types, and pixel-wise ground truth. Experiments on this benchmark dataset demonstrate that the proposed method outperforms the state-of-the-art salient object detection methods both qualitatively and quantitatively.
\end{abstract}

\begin{IEEEkeywords}
Salient object detection, optical remote sensing images, two-stream pyramid module, nested connections.
\end{IEEEkeywords}

\IEEEpeerreviewmaketitle

\section{Introduction}

\IEEEPARstart{S}{ALIENT} object detection aims at locating the most attractive and visually distinctive objects or regions from an image, which has been used in image/video segmentation \cite{R1}, image retargeting \cite{R2}, image foreground annotation \cite{R3}, thumbnail creation  \cite{R4}, image quality assessment \cite{R5}, and video summarization \cite{R6}. The last decades have witnessed the remarkable progress of saliency detection for nature scene image, especially including the deep learning-based methods with highly competitive performance \cite{RERVIEW}. In this paper, the nature scene image (usually RGB format) refers to the one captured by hand-held cameras or cameras mounted to objects on the ground, where the objects are typically in an upright orientation. It is worth mentioning that salient object detection is different from ordinary object detection or anomaly detection. First, object detection is a generalized task that aims to detect all the objects, while salient object detection only focuses on discovering the salient objects, and anomaly detection devotes to determining the abnormal objects. Second, object detection and anomaly detection always use bounding boxes to delineate the objects, while salient object detection generates a pixel-level saliency probability map.

Similar to the related works \cite{add1,rs4_add,add2,rs0,add3}, the RSIs used in this paper were collected from Google Earth with the spatial resolution ranging from 0.5m to 2m. Compared with the objects in nature scene images, the objects in RSIs (such as airports, buildings, and ships) usually have many different orientations, scales, and types since the RSIs are taken overhead \cite{add4,add5}. In addition, the optical RSI used in this paper is different from the hyperspectral image that includes more spectral bands information. The optical RSI used in this paper has the human eye friendly color presentation \cite{add1}. Facing the large-scale optical RSIs, people mainly focus on some local salient regions, \eg, man-made target and river system. Therefore, salient object detection in optical RSIs has extremely practical application values. \par

\begin{figure}[!t]
  \centering
\begin{minipage}[b]{0.19\linewidth}
  \centering
  \centerline{\includegraphics[width=1.1\linewidth,height=2.0\linewidth]{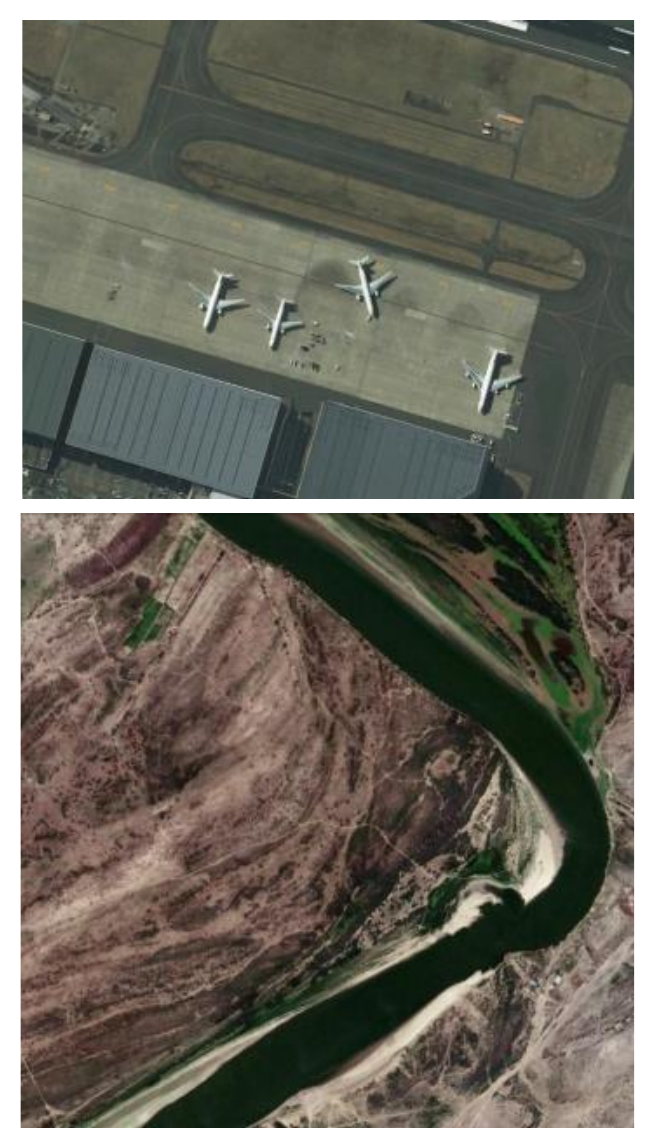}}
  \centerline{{\footnotesize (a)}}\smallskip
\end{minipage}
\begin{minipage}[b]{0.19\linewidth}
  \centering
  \centerline{\includegraphics[width=1.1\linewidth,height=2.0\linewidth]{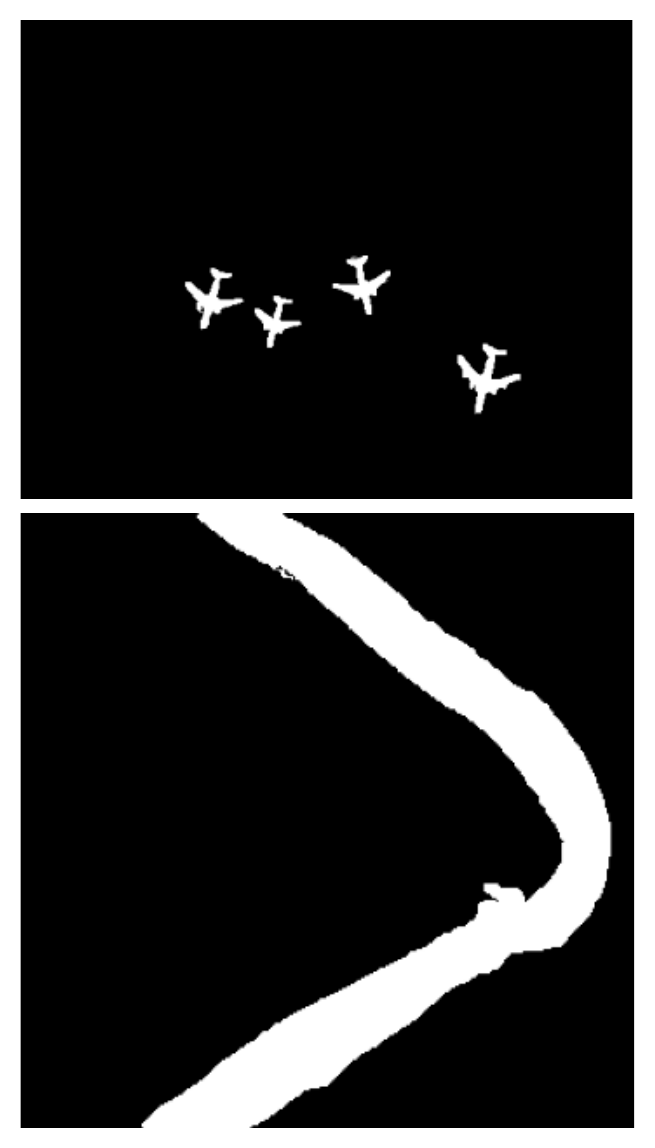}}
   \centerline{{\footnotesize (b)}}\smallskip
\end{minipage}
\begin{minipage}[b]{0.19\linewidth}
  \centering
  \centerline{\includegraphics[width=1.1\linewidth,height=2.0\linewidth]{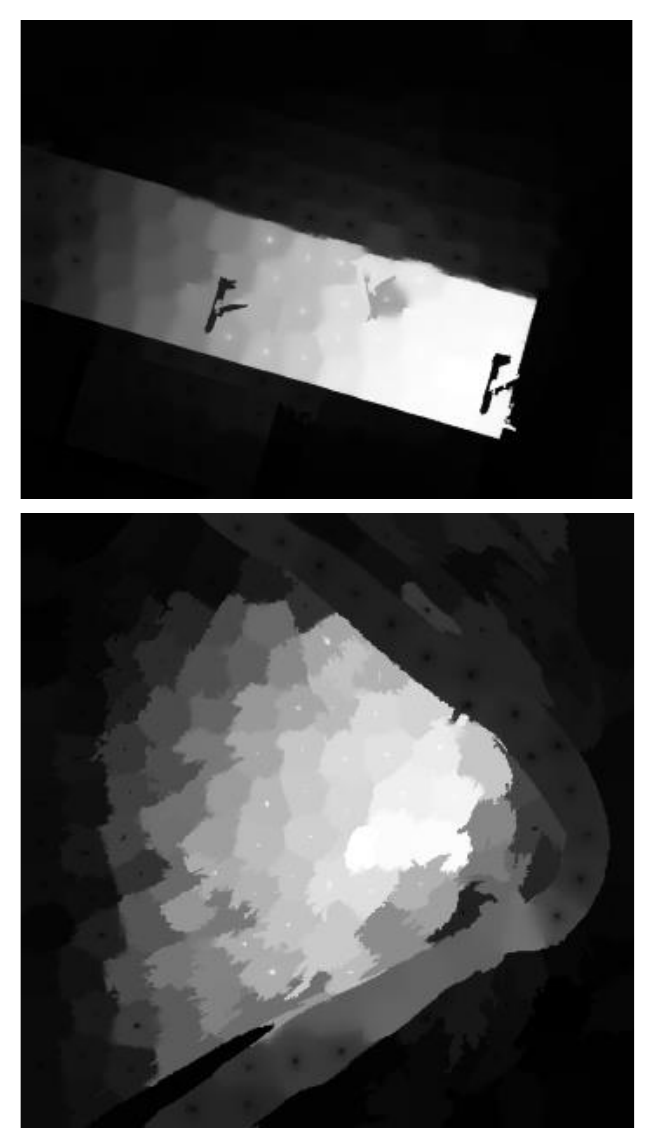}}
  \centerline{{\footnotesize (c)}}\smallskip
\end{minipage}
 \begin{minipage}[b]{0.19\linewidth}
  \centering
  \centerline{\includegraphics[width=1.1\linewidth,height=2.0\linewidth]{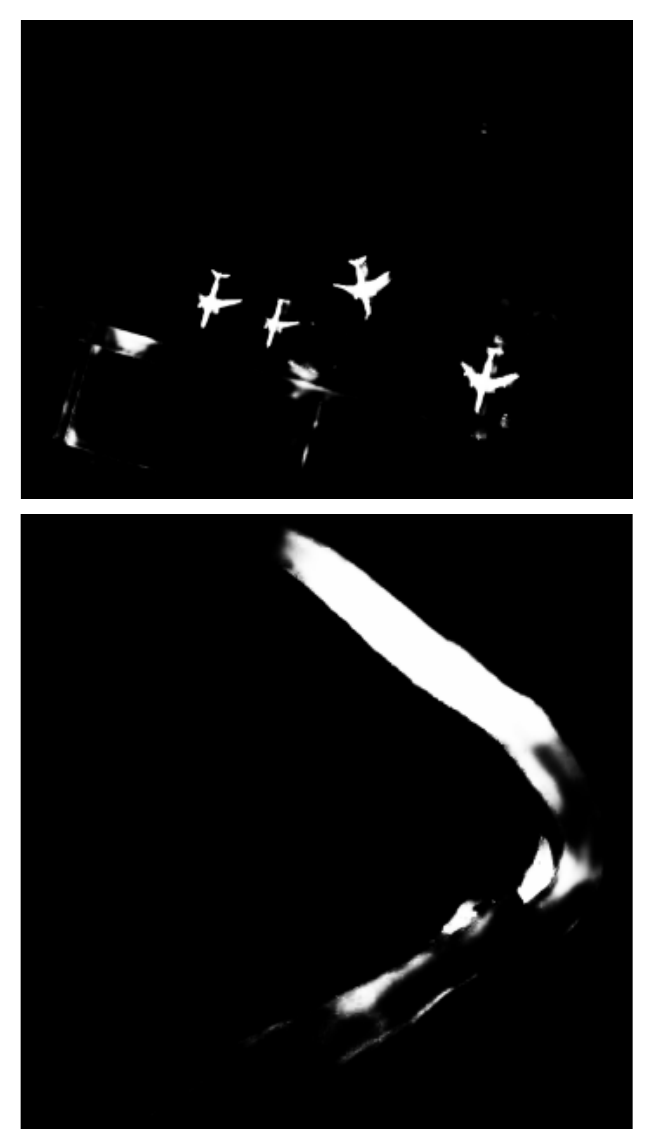}}
  \centerline{{\footnotesize (d)}}\smallskip
\end{minipage}
\begin{minipage}[b]{0.19\linewidth}
  \centering
  \centerline{\includegraphics[width=1.1\linewidth,height=2.0\linewidth]{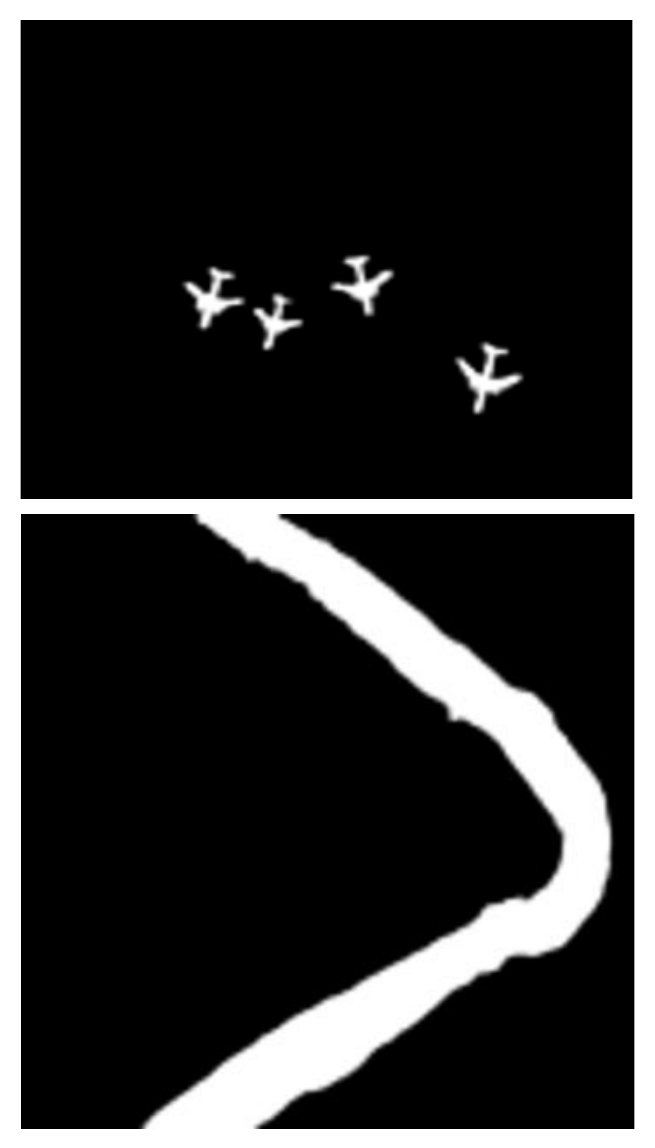}}
   \centerline{{\footnotesize (e)}}\smallskip
\end{minipage}
\caption{Visual results of salient object detection in optical RSIs by using different methods. (a) Optical RSIs. (b) Ground truth (GT). (c) RCRR \cite{RCRR}. (d) R3Net \cite{R3Net}. (e) Proposed LV-Net.}
\label{fig:1}
\end{figure}

Optical RSI is usually photographed outdoors in a high angle shot via the satellite and aerial sensors, and thus, there may be diversely scaled objects, various scenes and object types, cluttered backgrounds, and shadow noises. Sometimes, there is even no salient region in a real outdoor scene, such as the desert, forest, and sea. Due to these unique imaging conditions and diverse scene patterns, it is difficult to achieve satisfactory performance by directly transplanting the nature scene image saliency detection methods into the optical RSI.  Some visual examples of different saliency detection methods for optical RSIs are shown in Fig. \ref{fig:1}, where RCRR \cite{RCRR} is an unsupervised salient object detection method for nature scene image, and R3Net \cite{R3Net} is a recent deep learning-based salient object detection method fine-tuned on our constructed optical RSI dataset for salient object detection. As visible, the unsupervised RCRR method thoroughly fails to detect the salient objects from the optical RSIs, and the deep learning-based R3Net method also cannot highlight the salient objects accurately and completely. In contrast, the results of the proposed LV-Net are closer to the GT. In addition, there is no publicly available optical RSI dataset with saliency annotation for performance evaluation of salient object detection methods in optical RSIs. Given all that, it is highly desirable to design a specialized method for salient object detection in optical RSIs as well as a comprehensive dataset for performance evaluation. \par

In this paper, we propose a novel convolutional neural networks (CNN) architecture for salient object detection in optical RSIs, named LV-Net. The main contributions are summarized as follows.

\begin{itemize}
\item An end-to-end network for salient object detection in optical RSIs is proposed, including a two-stream pyramid module (L-shaped module) and an encoder-decoder module with nested connections (V-shaped module), which generalizes well to varying scenes and object patterns.

\item The L-shaped module learns a set of complementary features to address the scale variability of salient objects and capture local details, and the V-shaped module automatically determines the discriminative features to suppress cluttered backgrounds and highlight salient objects.

\item A challenging optical RSI dataset for salient object detection is constructed, including $800$ images with the corresponding pixel-wise ground truth. This dataset has been released for non-commercial use only. Moreover, the proposed method achieves the best performance against fourteen state-of-the-art salient object detection methods.

\end{itemize}

The rest of the paper is organized as follows. The related works on salient object detection are introduced in Section \ref{sec2}. Section \ref{sec3} presents the details of the proposed nested network with two-stream pyramid for salient object detection. In Section \ref{sec4}, the benchmark dataset and evaluation metrics, training strategies and implementation details, and experimental comparisons and analyses are discussed. Finally, the conclusion is drawn in Section \ref{sec6}.

\section{Related Work} \label{sec2}

The past decade has witnessed significant advances and performance improvements in salient object detection. In this section, we briefly review bottom-up and top-down saliency models for nature scene image, and then discuss the salient object detection in optical RSI. \par

\subsection{Saliency Detection in Nature Scene Image}
Bottom-up saliency detection model is stimulus-driven that aims at exploring low-level vision features. On one hand, some visual priors have been utilized to describe the properties of salient object based on the visual inspirations from the human visual system, such as contrast prior, background prior, and compactness prior. Cheng \emph{et al.} \cite{RC} proposed a simple and efficient salient object detection method based on global contrast, in which the saliency is defined as the color contrast to all other regions in the image. Zhu \emph{et al.} \cite{RBD} proposed a boundary connectivity measure to evaluate the background probability of a region. Then, a principled optimization framework integrating multiple low-level cues is used to achieve saliency detection. Zhou \emph{et al.} \cite{DCLC} combined the compactness prior with local contrast to discover salient object. Moreover, the saliency information is propagated on a graph through the diffusion framework. On the other hand, some traditional techniques have been introduced to achieve saliency detection, such as random walks, sparse representation, and matrix decomposition. Yuan \emph{et al.} \cite{RCRR} proposed a regularized random walk ranking model, which introduces prior saliency estimation to every pixel by taking both region and pixel image features into consideration. Li \emph{et al.} \cite{DSR} used the reconstruction error to measure the saliency of a region, where the salient region corresponds to a larger reconstruction error. Peng \emph{et al.} \cite{SMD} proposed a structured matrix decomposition method guided by high-level priors with two structural regularizations to achieve saliency detection.

\begin{figure*}[!t]
\centering
\centerline{\includegraphics[width=0.85\linewidth]{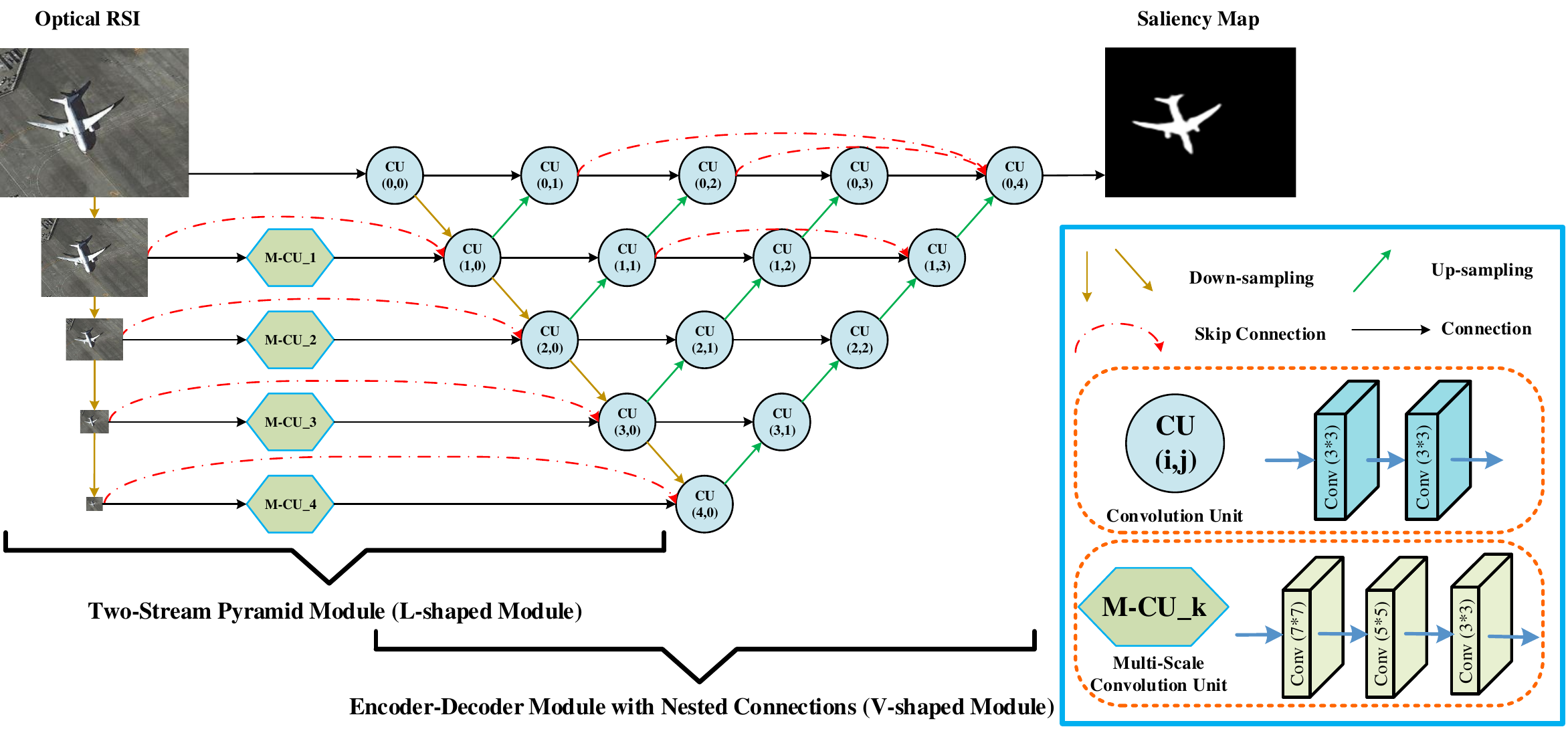}}
\caption{The architecture of the proposed LV-Net. LV-Net consists of two modules: 1) a two-stream pyramid module to deal with the scale variability of salient objects and capture local details, and 2) an encoder-decoder module with nested connections to gradually integrate encoder detail features and decoder semantical features across different scales. \textbf{Down-sampling} is implemented by $2 \times$ max pooling while \textbf{Up-sampling} is implemented by deconvolutional operation (transposed convolution) with kernels of size $3 \times 3$ and stride $2$. \textbf{ Convolution Unit (CU)} includes two consecutively convolutional operations with kernels of size $3 \times 3$ and stride $2$, where $i$ indexes the down-sampling layer along the encoder pathway, and $j$ indexes the convolutional layer of nested connections along the skip pathway. \textbf{Multi-Scale Convolution Unit (M-CU)} includes three consecutively convolutional operations with kernels of sizes $7\times 7$, $5\times 5$ and $3\times 3$, and stride $2$ respectively, where $k$ indexes the down-sampling layer along the input image. The convolutional layers used in the proposed LV-Net are followed by the ReLU activation function, expect for the CU$_{(0,4)}$ (\ie, followed by the sigmoid activation function). The convolutional layers in each CU and M-CU share the same number of output feature maps.}
\label{fig2}
\end{figure*}

Top-down saliency detection model is task-driven that entails supervised learning with ground truth. Especially, deep learning has demonstrated to be powerful for salient object detection. In recent years, numerous efforts have been made to design effective network architectures for extracting useful features which can characterize the salient objects \cite{ R3Net,SuperCNN, DCL, DSS, RADF, UCF, DSCLRCN,RFCN}. Deng \emph{et al.} \cite{R3Net} proposed a recurrent residual refinement network for saliency detection, where residual refinement blocks are leveraged to recurrently learn the difference between the coarse saliency map and the ground truth. Li and Yu \cite{DCL} proposed a deep contrast network for saliency detection, where the multi-scale fully convolutional stream captures the visual contrast saliency, and the segment-wise spatial pooling stream simulates saliency discontinuities along object boundaries. Hou \emph{et al.} \cite{DSS} introduced short connections into the skip-layer structures within the Holisitcally-nested Edge Detector (HED) architecture to achieve image saliency detection, which combines the low-level and high-level features at multiple scales. Hu \emph{et al.} \cite{RADF} introduced the recurrently aggregated deep features (RADF) into an FCNN to achieve saliency detection by fully exploiting the complementary saliency information captured in different layers. Zhang \emph{et al.} \cite{UCF} utilized an encoder Fully Convolutional Network (FCN) and a corresponding decoder FCN to detect salient object, in which a Reformulated dropout (R-dropout) is introduced to construct an uncertain ensemble of internal feature units. \par

\subsection{Saliency Detection in Optical RSI}
Compared to the saliency detection for nature scene image, only a small amount of work focuses on saliency detection for optical RSI. It is worth pointing out that although some so-called saliency detection methods for optical RSI have been proposed, most of them aim to realize other optical RSI processing tasks, such as Region-of-Interest (ROI) extraction and generalized object detection, by employing existing simple saliency models. Zhao \emph{et al.} \cite{rs5} proposed a sparsity-guided saliency detection method for optical RSIs, where the sparse representation is used to obtain global and background cues for saliency map integration. The authors collected some optical RSIs with the corresponding pixel-level saliency masks. Unfortunately, they did not make it publicly available.

Dong \emph{et al.} \cite{rs4_add} employed the visual saliency detection to locate the ROIs and homogeneous backgrounds in the prescreening state, which significantly reduces the false alarms. Then, a rotation-invariant descriptor was proposed for ship detection in optical RSIs. Li \emph{et al.} \cite{rs1} calculated the saliency in order to assist building extraction by combining the region contrast, boundary connectivity, and background constraints. In \cite{rs2}, the texture saliency and color saliency were integrated into the pixel-level saliency map to extract the ROI. Li \emph{et al.} \cite{rs3} proposed a hierarchical ROI detection method for optical RSIs, where the multilevel color histogram contrast is used to compute the saliency and obtain the preliminary regions. In \cite{rs4}, a two-way saliency model that combines vision-oriented saliency and knowledge-oriented saliency was proposed to estimate the airport position.

The unique imaging conditions and diverse scene patterns pose new challenges for salient object detection in optical RSIs. Thus, it is difficult to achieve satisfactory performance by directly using the existing nature scene image saliency detection methods. Moreover, due to the lack of sufficient training data and elaborate network architecture to handle the  multiple salient objects with diverse scales and complicated backgrounds, the superiority of deep learning has not been demonstrated in the salient object detection of optical RSIs. To address the above-mentioned problems, we propose a deep learning-based salient object detection method specially designed for optical RSIs, and construct the first publicly available optical RSI dataset for salient object detection.\par

\section{Proposed Method}\label{sec3}

\subsection{Framework}

In Fig. \ref{fig2}, we present the LV-Net network architecture including a two-stream pyramid module (L-shaped module) and an encoder-decoder module with nested connections (V-shaped module), which takes an optical RSI as input and outputs its saliency map.

To address the different scales of salient objects in optical RSIs, an L-shaped module is designed in the LV-Net. First, we progressively down-sample the input optical RSI for input pyramid generation. Then, we extract the multi-scale feature representations of each down-sampled input through a multi-scale convolution unit, and finally form a multi-scale feature pyramid. The input pyramid preserves original detail features of input images, and the feature pyramid provides abstract semantic features. Both the detail and semantic features are significant for the task of salient object detection. Therefore, we concatenate multi-resolution input versions and multi-scale features at different levels to form the two-stream pyramid and obtain complementary features.

\begin{figure}[!t]
\centering
\centerline{\includegraphics[width=0.8\linewidth]{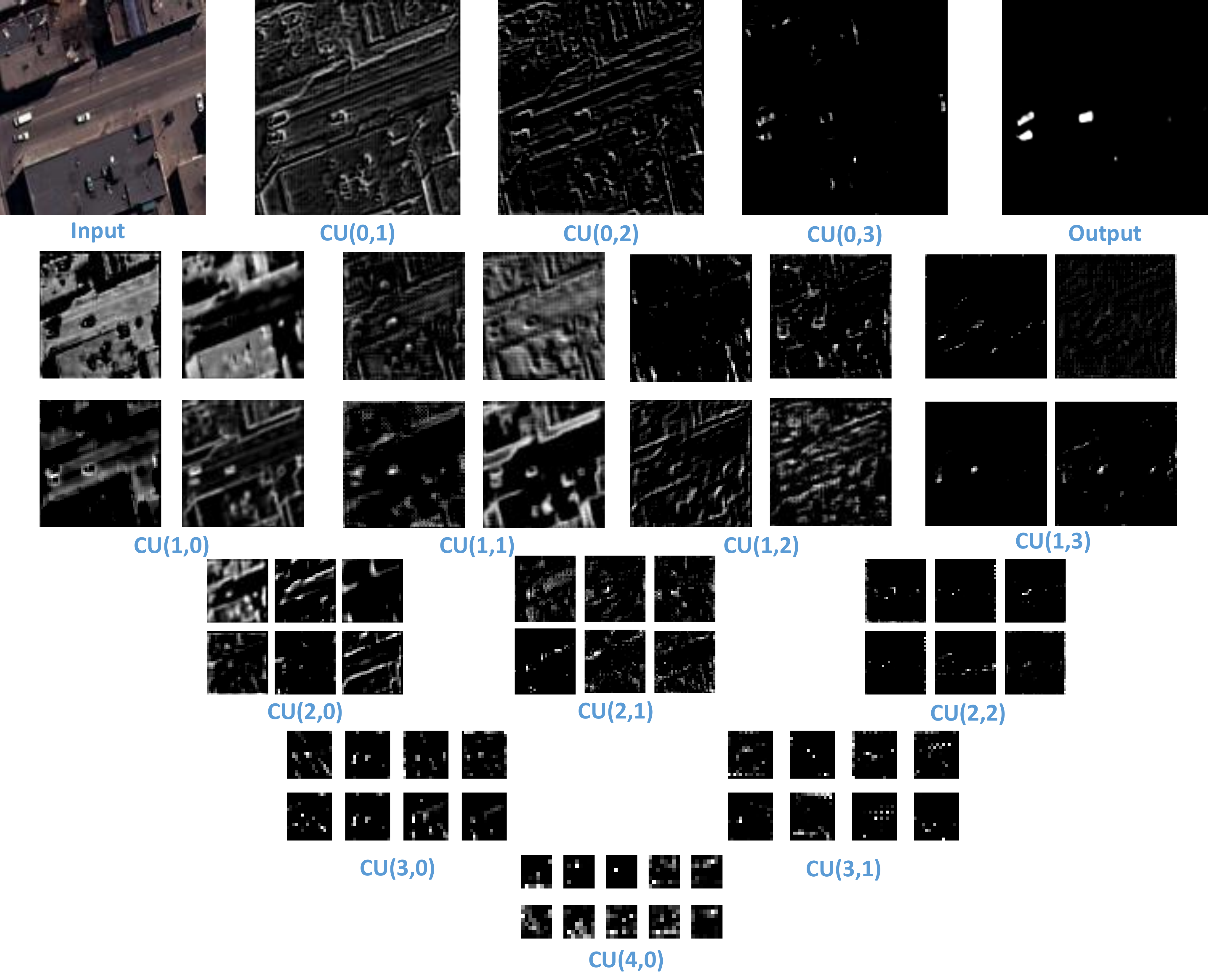}}
\caption{Features visualization of the proposed LV-Net. For the limited space, we only present several feature maps.}
\label{fig4}
\end{figure}

The L-shaped module can extract detail features and semantic features, but it is still not enough to accurately and completely detect the salient objects in optical RSIs. Thus, the complementary features hierarchically extracted by the two-stream pyramid structure are passed to an encoder-decoder module, which gradually integrates encoder detail features and decoder semantic features with nested connections. At the end, the salient regions of an input optical RSI are predicted by the integrated features in a deeply supervised manner. From the view of feature type, by combining the features from L-shaped and V-shaped modules, the final features are relatively more comprehensive than the features only from the L-shaped module or the features only from the V-shaped module. From the view of network optimization, the purpose of networks is to learn the discriminative feature representations to assist saliency detection and boost the final saliency performance. In Fig. \ref{fig4}, we provide the features visualization of the proposed network. As visible, the features progressively become discriminative (close to the final saliency map) which can effectively distinguish the foreground and background, such as the features in CU(0,3) and CU(1,3). In addition, one can find that the detail features (\emph{e.g.}, edges and textures) in the encoder path become more and more abstract with the down-sampling, while the cluttered and noisy backgrounds gradually vanish with the nested connections and up-sampling in the decoder path. Thus, from different views, the features extracted by the combination of the L-shaped and V-shaped modules are comprehensive and discriminative. Next, we will illustrate the advantages and implementation details of these two modules.

\subsection{Two-Stream Pyramid Module}
As mentioned earlier, the type and scale of the objects in the optical RSI are variable and diverse, including some small scaled airplanes or large bodies of water. To deal with the scale variability of image patterns, we design an input pyramid structure and pass scaled versions through our network.

\begin{figure}[!t]
\centering
\centerline{\includegraphics[width=1\linewidth,height=0.2\linewidth]{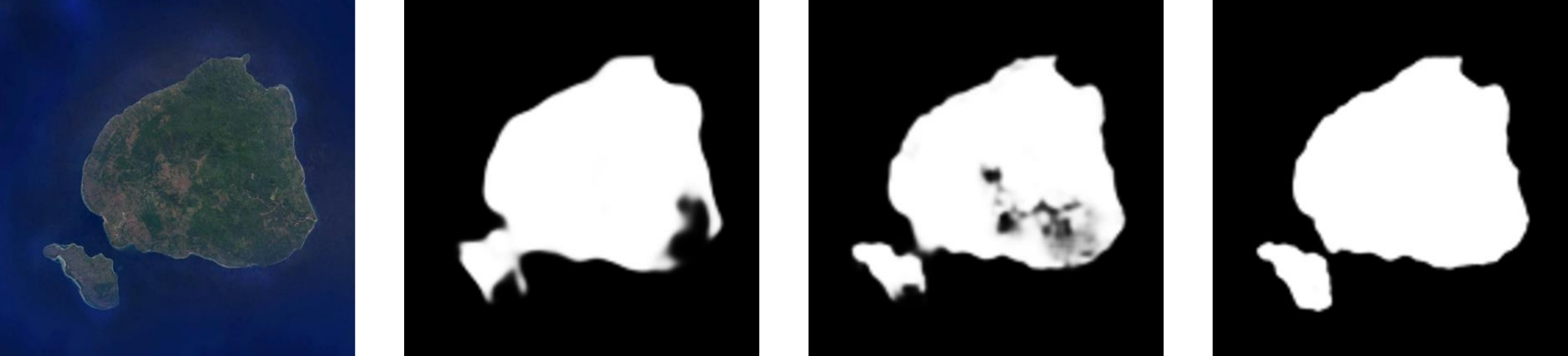}}
\caption{The saliency maps obtained by different networks. From left to right are input image, the saliency map obtained by the LV-Net without feature pyramid, the saliency map obtained by the LV-Net without input pyramid, the saliency map obtained by our LV-Net with two-stream pyramid.}
\label{fig_complementary}
\end{figure}

We employ the $2 \times$ max pooling layer to progressively down-sample input image and generate the input pyramid:
\begin{equation}
\label{equ_1}
I_{ds}^{k}=maxpool(I_{in}),
\end{equation}
where $k\in\{1,2,3,4\}$ indexes the down-sampling layer along the input image, $I_{ds}^{k}$ is the $k^{th}$ 2$\times$ downsampling result of the input $I_{in}$, and $maxpool$ represents the max pooling operation.

Then, we extract multi-scale features from the scaled versions by the M-CU to form the multi-scale feature pyramid. The number of output feature maps of each convolutional layer in the M-CU is denoted as $32\times2^k$. Here, $k$ indexes the down-sampling layer along the input image. The M-CU operation can be expressed as:
\begin{equation}
\label{equ_2}
F_{7\times7}^{k}=\sigma (\mathbf{W}_{7\times7}^{k}*I_{ds}^{k}+\mathbf{b}_{7\times7}^{k}),
\end{equation}
\begin{equation}
\label{equ_3}
F_{5\times5}^{k}=\sigma (\mathbf{W}_{5\times5}^{k}*F_{7\times7}^{k}+\mathbf{b}_{5\times5}^{k}),
\end{equation}
\begin{equation}
\label{equ_4}
F_{3\times3}^{k}=\sigma (\mathbf{W}_{3\times3}^{k}*F_{5\times5}^{k}+\mathbf{b}_{3\times3}^{k}),
\end{equation}
where $F_{7\times7}^{k}$ is a set of features extracted from $I_{ds}^{k}$, $F_{5\times5}^{k}$ is a set of features extracted from $F_{7\times7}^{k}$, $F_{3\times3}^{k}$ is a set of features extracted from $F_{5\times5}^{k}$, $*$ represents a convolutional operation, $\mathbf{W}$ and $\mathbf{b}$ stand for the weight and bias, and $\sigma$ is the element-wise rectified linear unit (ReLU) activation function \cite{ReLU}.

Finally, we concatenate multi-resolution input versions and multi-scale features at different levels to form the two-stream pyramid. The main reasons for using the two-stream pyramid lie in two aspects: (1) The input pyramid includes several scaled input images, which preserves original detail features of input images, but lacks semantic information. In contrast, the feature pyramid extracts multi-level features by consecutive convolution operations, which provides abstract semantic features, but lacks detail information. However, both detail features and semantic features are significant for the task of salient object detection. Therefore, we concatenate multi-resolution input versions and multi-scale features at different levels to form the two-stream pyramid to obtain complementary feature representations. Here, from the view of feature content, the concrete detail features and abstract semantic features are complementary. In Fig. \ref{fig_complementary}, we present a visual example to explain the complementary behavior of the features extracted by the two-stream pyramid. For the LV-Net without the feature pyramid, the final saliency map may be incomplete since the semantic features extracted by the feature pyramid are removed. For the LV-Net without the input pyramid, the final saliency map may lose some details, such as the holes in the saliency map. In contrast, when the input pyramid and the feature pyramid are combined, the final saliency map preserves complete structures and clear details. (2) The two-stream pyramid can attain multiple level receptive fields by the consecutive convolution operations and the preservation of original detail features. Thus, we did not use the time-consuming dense connection pattern like DenseNet \cite{DenseNet} to stack the features from all layers. We will demonstrate the effects of two-stream pyramid in Section \ref{sec4}.

\subsection{Encoder-Decoder Module with Nested Connections}
Essentially, our architecture is a deeply supervised encoder-decoder network, where the encoder and decoder pathways are connected through a series of nested connections. The basic idea behind the use of nested connections is that the nested connections would automatically select more discriminative saliency features by the supervised learning, so that it could facilitate the fusion of encoder-decoder features and remit the interferences of cluttered and noisy backgrounds. In an encoder-decoder system, the model that only uses the high-level decoder features is unable to capture the details of the salient objects, while the model with only the low-level encoder features fails to distinguish the salient objects from the complicated backgrounds. To accurately capture the salient objects with exact boundaries, some encoder-decoder network architectures (\eg, U-Net \cite{U-Net}) usually concatenate encoder detail features and decoder semantical features through the brute-force skip connections, whose effectiveness has been proven in the saliency detection for simple nature scene image and medical image segmentation. Unfortunately, we found that the brute-force skip connections can degrade the quality of saliency prediction because the cluttered and noisy encoder features can also be passed through the prediction layer, especially for optical RSIs with complicated backgrounds. The `bad' features seriously affect the accuracy of the saliency prediction. Therefore, we use the nested connections to gradually filter out the `bad' distractive features and make salient objects stand out by task-driven learning. We will conduct an ablation study to verify our findings in Section \ref{sec4}.

Specifically, the operations in the encoder-decoder module with nested connections can be expressed as follows.

\begin{equation}
\label{equ_5}
F_{(0,0)}={\rm CU}_{(0,0)}(I_{in}),
\end{equation}

\begin{equation}
\label{equ_6}
F_{(p,0)}={\rm CU}_{(p,0)}(\{F_{3\times3}^{p},I_{ds}^{p},F_{(p-1,0)}^{down} \}),
\end{equation}
where $F_{(0,0)}$ is a set of features extracted by convolution unit CU$_{(0,0)}$, $F_{(p,0)}$ is a set of features extracted by CU$_{(p,0)}$, $p\in\{1,2,3,4\}$, $F_{(p-1,0)}$ is a set of features extracted by CU$_{(p-1,0)}$, $F_{(p-1,0)}^{down}$ is the $2 \times$ down-sampling result of $F_{(p-1,0)}$, $F_{3\times3}^{p}$ is a set of outputs of $p^{th}$ M-CU, $I_{ds}^{p}$ is the $p^{th}$ $2\times$ down-sampling result of the input optical RSI $I_{in}$, \{$F_{3\times3}^{p},I_{ds}^{p},F_{(p-1,0)}$\} represents the concatenation of $F_{3\times3}^{p}$, $I_{ds}^{p}$ and $F_{(p-1,0)}^{down}$.  Note that, $\{\cdot\}$ indicates the feature concatenation operation (not the element-wise addition).
\begin{equation}
\label{equ_7}
F_{(q,1)}={\rm CU}_{(q,1)}(\{F_{(q,0)},F_{(q+1,0)}^{up}\}),
\end{equation}
\begin{equation}
\label{equ_8}
F_{(m,2)}={\rm CU}_{(m,2)}(\{F_{(m,1)},F_{(m+1,1)}^{up}\}),
\end{equation}
where $F_{(q,1)}$ is a set of features extracted by the convolution unit CU$_{(q,1)}$, and $q\in\{0,1,2,3\}$. $F_{(m,2)}$ is a set of features extracted by the convolution unit CU$_{(m,2)}$, $m\in\{0,1,2\}$. $F_{(q+1,0)}^{up}$ is the up-sampling result of $F_{(q+1,0)}$, $F_{(m+1,1)}^{up}$ is the up-sampling result of $F_{(m+1,1)}$.
\begin{equation}
\label{equ_9}
F_{(0,3)}={\rm CU}_{(m,2)}(\{F_{(0,2)},F_{(1,2)}^{up}\}),
\end{equation}
\begin{equation}
\label{equ_10}
F_{(1,3)}={\rm CU}_{(m,2)}(\{F_{(1,2)},F_{(2,2)}^{up}, F_{(1,1)}\}),
\end{equation}
where $F_{(0,3)}$ and  $F_{(1,3)}$  are the features extracted by the convolution units CU$_{(0,3)}$ and CU$_{(1,3)}$, respectively.
\begin{equation}
\label{equ_11}
F_{(0,4)}= {\rm CU}_{(0,4)}(\{F_{(0,3)},F_{(1,3)}^{up},F_{(0,1)},F_{(0,2)}\}),
\end{equation}
where $F_{(0,4)}$ is the output of the CU$_{(0,4)}$, and also is the final saliency map. The number of output feature maps of each CU in the encoder-decoder module is denoted as $64 \times2^i$, except for the CU$_{(0,4)}$ (\ie, the output number is 1), where $i$ indexes the down-sampling layer along the encoder pathway.

\subsection{Loss Function}
Following \cite{DSS,ELD-Net}, we also minimize the sigmoid cross-entropy loss to learn the saliency prediction mapping:

\begin{equation}
\label{equ_1}
L=-(ylog(z)+(1-y)log(1-z)),
\end{equation}
where $0$ and $1$ represent the non-salient and salient region labels, $y$ and $z$ are the true label and the score of the predicted result. However, we found that this loss function does not always work ($L \to \infty$) when the predicted score $z$  is $0$ or $1$. It is possible for optical RSI when there is no salient object. Thus, we rewrite the sigmoid cross-entropy loss as:
\begin{equation}
\label{equ_2}
L=-(ylog(F_{clip}(z))+(1-y)log(1-F_{clip}(z))).
\end{equation}
where $F_{clip}$ is a function that returns a tensor of the same type and shape as input with its values clipped to $\rho$ and $\mu$. Specifically, any values less than $\rho$ are set to $\rho$, while any values greater than $\mu$ are set to $\mu$. Here, we set $\rho$ and $\mu$ to $1e^{-15}$ and $1-1e^{-15}$, respectively.

\section{Experiments}\label{sec4}

In this section, we first introduce the constructed benchmark dataset for salient object detection in optical RSI, and then illustrate the evaluation metrics, training strategies, and implementation details. Then, we compare the proposed LV-Net with state-of-the-art methods to demonstrate its advantage. At last, some ablation studies are discussed to verify the role of each component and analyze the effects of parameter settings.

\subsection{Benchmark Dataset}

\begin{figure}[!t]
\centering
\centerline{\includegraphics[width=1\linewidth,height=0.25\linewidth]{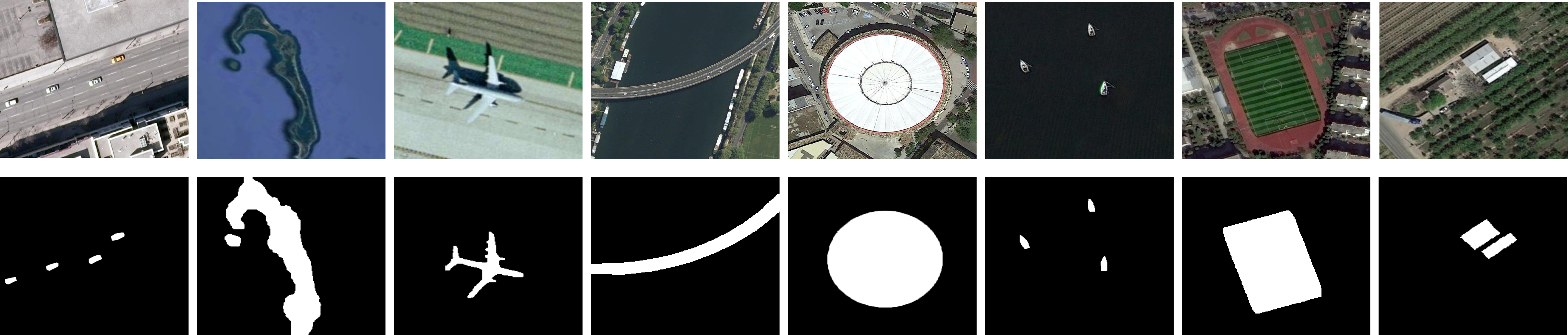}}
\caption{Sample images from the constructed ORSSD dataset. The first row are the optical RSIs. The second row are the pixel-wise annotation.}
\label{fig_add}
\end{figure}

To the best of our knowledge, there is no publicly available optical RSI dataset for salient object detection. To fill the gap, we collected $800$ optical RSIs to construct a dataset for salient object detection, named ORSSD dataset, and the manually pixel-wise annotation for each image is provided. Most of the original optical RSIs are collected from the Google Earth, and several RSIs are from the existing RSI databases, such as NWPU VHR-10 dataset \cite{rs0_1}, AID dataset \cite{AID}, LEVIR dataset \cite{LEVIR}, and NWPU-RESISC45 dataset\cite{rs0_3}. For ground truth annotation, we firstly selected 5 people with relevant background to mark the salient objects independently. Then, only the objects marked four times would be selected as the ground truth of salient objects. The ORSSD dataset is very challenging, because a) the spatial resolution is diverse, such as $1264\times 987$, $800\times 600$, and $256\times 256$, b) the background is cluttered and complicated, including some shadows, trees, and buildings, c) the type of salient objects is various, including airplane, ship, car, river, pond, bridge, stadium, beach, \etc, and d) the number and size of salient objects are variable, even in some scenes there are no salient object, such as the desert and thick forest. Some of the sample images in the constructed ORSSD dataset are shown in Fig.~\ref{fig_add}. The ORSSD dataset can be used for deep network training and the performance evaluation of salient object detection in optical RSIs, which is available from our project \url{https://li-chongyi.github.io/proj_optical_saliency.html}.\par

\subsection{Evaluation Metrics}
\begin{figure*}[!t]
\centering
\centerline{\includegraphics[width=0.8\linewidth]{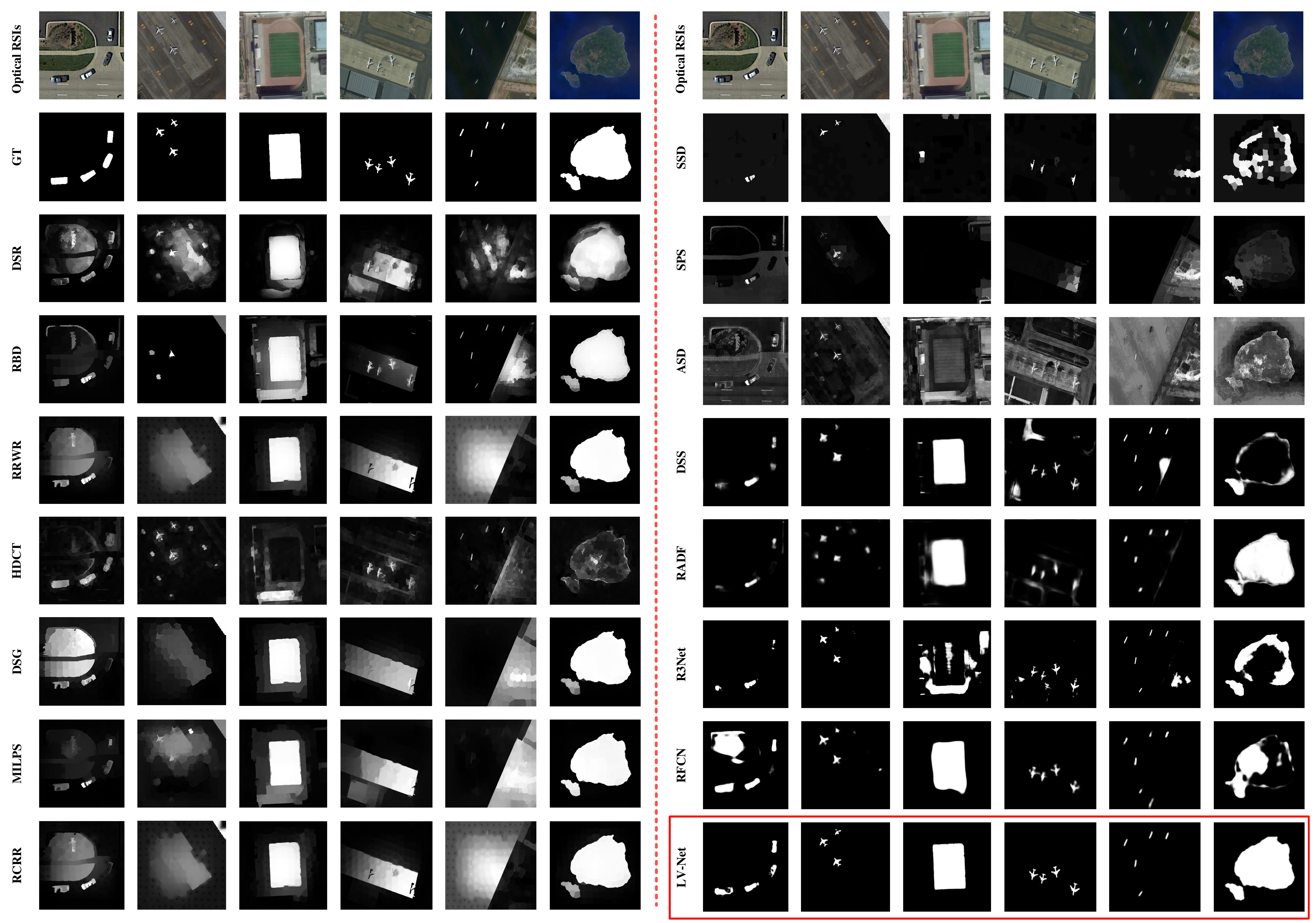}}
\caption{Visual comparisons of different methods. Our results are marked by the red box.}
\label{fig5}
\end{figure*}

To quantitatively evaluate the performance of different methods, the Precision-Recall (PR) curve, F-measure, MAE score, and S-measure are employed.

Using a series of fixed integers from $0$ to $255$, a saliency map can be thresholded as some binary saliency masks. Then, comparing the binary mask with the ground truth, the precision and recall scores are obtained. The PR curve is drawn under different combination of precision and recall scores, where the vertical axis denotes the precision score, and the horizontal axis corresponds to the recall score. The closer the PR curve is to the coordinates $(1,1)$, the better the performance achieves.

As a comprehensive measurement, F-measure is defined as the weighted harmonic mean of precision and recall \cite{Fmeasure2}, \emph{i.e.}, $F_{\beta}=\frac{(1+\beta^{2})Precision\times Recall}{\beta^{2}\times Precision+ Recall}$, where $\beta^{2}$ is set to $0.3$ for emphasising the precision as suggested in \cite{Fmeasure,Fmeasure2}. The larger $F_{\beta}$ value indicates the better comprehensive performance.

MAE score \cite{MAE} calculates the difference between the continuous saliency map $S$ and ground truth $G$, \emph{i.e.}, $MAE =\frac{1}{w\times h} \sum_{i=1}^{w} \sum_{j=1}^{h} |S(i,j)-G(i,j)|$, where $w$ and $h$ represent the width and height of the image, respectively. The smaller the MAE score is, the more similar to the ground truth, and the better performance achieves.

S-measure \cite{S-measure} evaluates the structural similarity between the saliency map and ground truth, \emph{i.e.}, $S_m = \alpha \times S_o+(1-\alpha) \times S_r$, where $\alpha$ is set to $0.5$ for assigning equal contribution to both region $S_r$ and object $S_o$ similarity as suggested in \cite{S-measure}. The larger $S_m$ value demonstrates better performance in terms of the structural similarity.\par

\subsection{Training Strategies and Implementation Details}

\textbf{Network Training.} We randomly selected $600$ images from ORSSD dataset for training and the rest $200$ images as the testing dataset. We augmented data with flipping and rotation, and then obtained seven additional augmented versions of the original training data. In the training phase, the samples were resized to size $128\times 128$ due to our limited memory. At last, the augmented training data provided $4800$ pairs of images.

\textbf{Implementation Details.} We implemented the proposed LV-Net with TensorFlow on a PC with an Intel(R) i$7$ $6700$ CPU, $32$GB RAM, and an NVIDIA GeForce GTX $1080$Ti GPU. During training, a batch-mode learning method with a batch size of $16$ was applied. The filter weights of each layer were initialized by Xavier policy \cite{Xavier}, and the bias was initialized as constant. We used ADAM \cite{Kingma2014} for network optimization and fixed the learning rate to $1e^{-4}$ in the entire training procedure. To illustrate the implementation details, we report the input size and output size for each convolutional unit in the procedure of network training in Table \ref{add1}.

\begin{table}[!t]
\renewcommand\arraystretch{1}
\caption{Input Size and Output Size (batch, width, height, channel) of Each Convolutional Unit in the Proposed LV-Net.}
\begin{center}
\setlength{\tabcolsep}{4mm}{
\begin{tabular}{c|c|c}
\hline
  & Input Size &  Output Size \\
\hline\hline
M-CU\_1& (16,64,64,3) & (16,64,64,64)  \\
\hline
M-CU\_2 & (16,32,32,3) & (16,32,32,128)  \\
\hline
M-CU\_3& (16,16,16,3) & (16,16,16,256)  \\
\hline
M-CU\_4& (16,8,8,3) & (16,8,8,512)  \\
\hline
CU$_{(0,0)}$& (16,128,128,3) & (16,128,128,64)  \\
\hline
CU$_{(1,0)}$&  (16,64,64,67)& (16,64,64,128)\\
\hline
CU$_{(2,0)}$&  (16,32,32,131) & (16,32,32,256 \\
\hline
CU$_{(3,0)}$&  (16,16,16,259) & (16,16,16,512 \\
\hline
CU$_{(4,0)}$&  (16,8,8,515) & (16,8,8,1024 \\
\hline
CU$_{(0,1)}$& (16,128,128,192) & (16,128,128,64) \\
\hline
CU$_{(1,1)}$& (16,64,64,384) & (16,64,64,128)\\
\hline
CU$_{(2,1)}$& (16,32,32,768) & (16,32,32,256)\\
\hline
CU$_{(3,1)}$& (16,16,16,1536) & (16,16,16,512)\\
\hline
CU$_{(0,2)}$& (16,128,128,192) & (16,128,128,64)\\
\hline
CU$_{(1,2)}$& (16,64,64,384) & (16,64,64,128)\\
\hline
CU$_{(2,2)}$& (16,32,32,768) & (16,32,32,256)\\
\hline
CU$_{(0,3)}$&  (16,128,128,192) & (16,128,128,64)\\
\hline
CU$_{(1,3)}$&  (16,64,64,768) & (16,64,64,128)\\
\hline
CU$_{(0,4)}$&  (16,128,128,320) & (16,128,128,1)\\
\hline
\end{tabular}}
\end{center}
\label{add1}
\end{table}

\subsection{Comparison with State-of-the-art Methods}

\begin{figure}[!t]
\centering
\centerline{\includegraphics[width=0.8\linewidth]{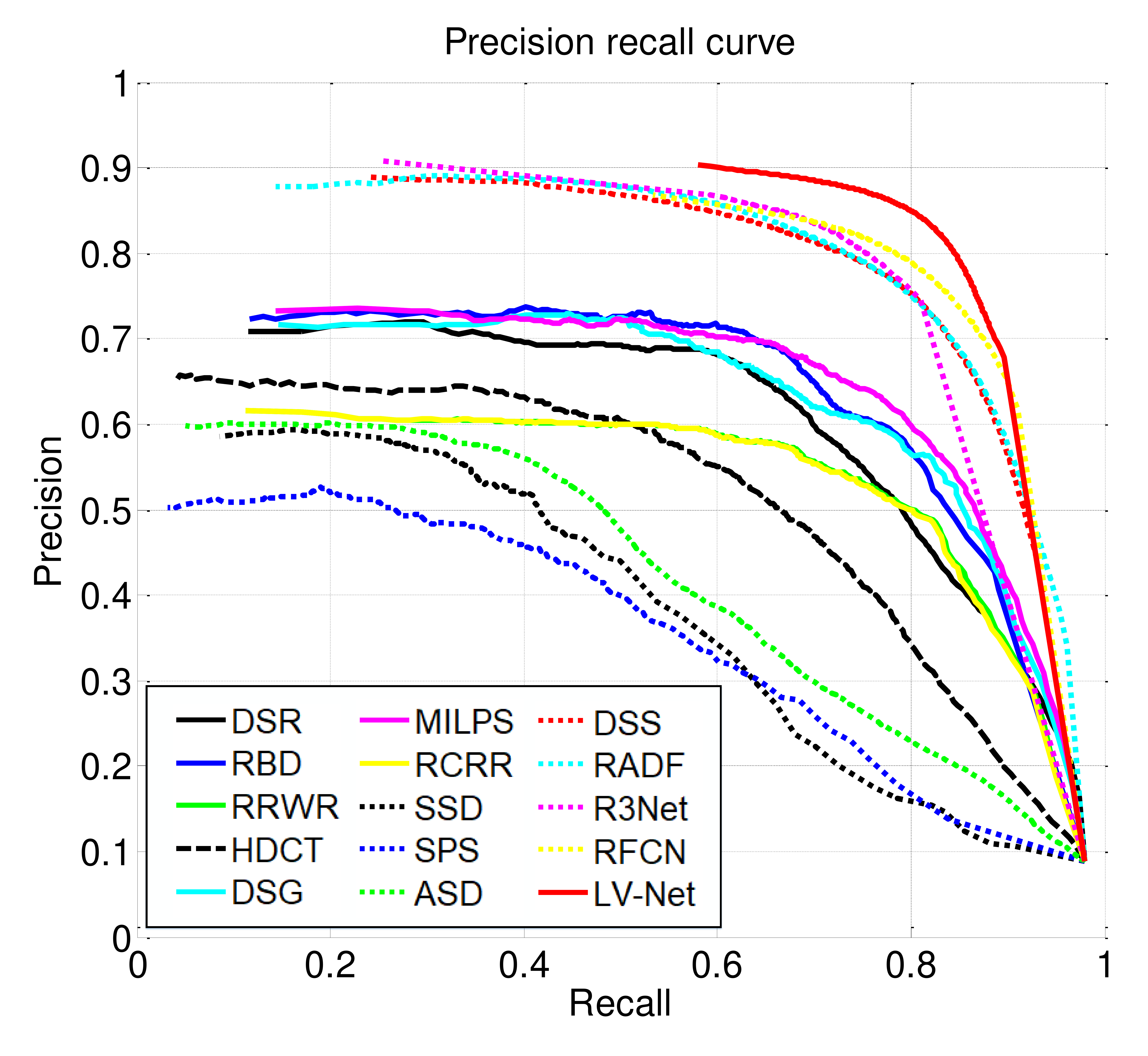}}
\caption{PR curves of different methods on the test subset of ORSSD dataset. }
\label{fig6}
\end{figure}

We compare the proposed method with fourteen state-of-the-art salient object detection methods on the testing subset of ORSSD dataset, including seven unsupervised methods (DSR \cite{DSR}, RBD \cite{RBD}, RRWR \cite{RRWR}, HDCT \cite{HDCT}, DSG \cite{DSG}, MILPS \cite{MILPS}, and RCRR \cite{RCRR}), four deep learning-based methods (DSS \cite{DSS}, RADF \cite{RADF}, R3Net \cite{R3Net}, and RFCN \cite{RFCN}), and three saliency detection methods in optical RSIs (SSD \cite{rs5}, SPS \cite{rs2}, and ASD \cite{rs4}). All the results are generated by the source codes or provided by the authors. For a fair comparison, we retrained the compared deep learning-based saliency detection methods using the same training data with the proposed LV-Net and the default parameter settings in the corresponding models. We tuned to generate best results of the compared methods. The visual comparisons are shown in Fig. \ref{fig5}, and the quantitative evaluations are reported in Fig. \ref{fig6} and Table \ref{tab1}. \par

In Fig. \ref{fig5}, six optical RSIs with different salient objects (\eg, ships, cars, airplanes, playground, and island) are illustrated. From it, we can see that the unsupervised methods (\ie, DSR, DSG, and RCRR) cannot highlight the salient objects effectively and completely. For example, in the first scene, some backgrounds (\eg, tree lawn) are wrongly detected as the salient regions, and the four cars could not be totally detected by these unsupervised methods. For the salient objects with small sizes (\eg, the airplanes in the second image and fourth image, and the ships in the fifth image), the unsupervised methods cannot handle this challenging situation effectively. Especially, the RCRR method \cite{RCRR} is completely powerless to locate these salient objects in such a challenging scene. In terms of the existing salient object detection methods for optical RSI (\eg, SSD \cite{rs5}, SPS \cite{rs2}, and ASD \cite{rs4}), they fail to locate the salient objects in the challenging optical RSIs because they only use or modify the hand-crafted features designed for nature scene images in a coarse way and ignore the unique characteristics of optical RSIs. By contrast, the deep learning-based methods achieve obvious superiority in performance. For example, in the first image, the complicated backgrounds are effectively suppressed by the deep learning-based methods, and most of the salient objects can be clearly detected by the DSS method \cite{DSS}. However, there are several drawbacks in these methods, including missing detection (\eg, the cars in the first image using the RADF method \cite{RADF}), wrong detection (\eg, the ships in the fifth image using DSS method \cite{DSS} and R3Net method \cite{R3Net}, and incomplete detection (\eg, the playground in the third image using the R3Net method \cite{R3Net}, and the island in the last image using the DSS \cite{DSS}). Benefiting from both the two-stream pyramid and nested connections, more comprehensive and discriminative representations can be learned to filter out the interferences in the complicated backgrounds, refine the details of the salient objects, and improve the identification accuracy of salient objects. As visible, the proposed LV-Net method highlights the salient object more accurate and complete. Moreover, the backgrounds by our results are suppressed effectively.\par

\begin{table}[!t]
\renewcommand\arraystretch{1.2}
\caption{Quantitative Comparisons with Different Methods on the Testing Subset of ORSSD Dataset.}
\begin{center}
\setlength{\tabcolsep}{2mm}{
\begin{tabular}{c|c|c|c|c|c}
\hline
Method & Precision & Recall & $F_{\beta}$ & MAE & $S_m$ \\
\hline\hline
DSR \cite{DSR} & $0.6829$ & $0.5972$ & $0.6610$ & $0.0859$ & $0.7082$ \\
\hline
RBD \cite{RBD} & $0.7080$ & $0.6268$ & $0.6874$ & $0.0626$ & $0.7662$ \\
\hline
RRWR \cite{RRWR} & $0.5782$ & $0.6591$ & $0.5950$ & $0.1324$ & $0.6835$ \\
\hline
HDCT \cite{HDCT} & $0.6071$ & $0.4969$ & $0.5775$ & $0.1309$ & $0.6197$ \\
\hline
DSG \cite{DSG} & $0.6843$ & $0.6007$ & $0.6630$ & $0.1041$ & $0.7195$ \\
\hline
MILPS \cite{MILPS} & $0.6954$ & $0.6549$ & $0.6856$ & $0.0913$ & $0.7361$ \\
\hline
RCRR \cite{RCRR} & $0.5782$ & $0.6552$ & $0.5944$ & $0.1277$ & $0.6849$ \\
\hline
SSD \cite{rs5} & $0.5188$ & $0.4066$ & $0.4878$ & $0.1126$ & $0.5838$ \\
\hline
SPS \cite{rs2} & $0.4539$ & $0.4154$ & $0.4444$ & $0.1232$ & $0.5758$ \\
\hline
ASD \cite{rs4} & $0.5582$ & $0.4049$ & $0.5133$ & $0.2119$ & $0.5477$ \\
\hline
DSS \cite{DSS} & $0.8125$ & $0.7014$ & $0.7838$ & $0.0363$ & $0.8262$ \\
\hline
RADF \cite{RADF} & $0.8311$ & $0.6724$ & $0.7881$ & $0.0382$ & $0.8259$  \\
\hline
R3Net \cite{R3Net} & $0.8386$ & $0.6932$ & $0.7998$ & $0.0399$ & $0.8141$ \\
\hline
RFCN \cite{RFCN} & $0.8239$ & $0.7376$ & $0.8023$ & $0.0293$ & $0.8437$  \\
\hline
LV-Net & $\textbf{0.8672}$ & $\textbf{0.7653}$ & $\textbf{0.8414}$ & $\textbf{0.0207}$ & $\textbf{0.8815 }$ \\
\hline
\end{tabular}}
\end{center}
\label{tab1}
\end{table}

The PR curves are shown in Fig. \ref{fig6}. The PR curve describes the different combination of precision and recall scores, and the closer the PR curve is to the coordinates $(1,1)$, the better performance achieves. Compared with other methods, the proposed LV-Net algorithm achieves a higher recall score while achieving a higher precision score, and thus, its PR curve is much higher than other methods with a large margin. In other words, the proposed LV-Net achieves a win-win situation between precision and recall, which demonstrates the effectiveness of the proposed algorithm. Table \ref{tab1} reports the quantitative measures of different methods, including the Precision score, Recall score, F-measure, MAE score, and S-measure. The best result for each evaluation is in bold. The same conclusion can be drawn from Table \ref{tab1}, that is, the performance of the proposed LV-Net is significantly superior to the state-of-the-art methods in terms of Precision score, Recall score, F-measure, MAE score, and S-measure. Compared with the \emph{\textbf{second best method}}, the performance improvement of the proposed LV-Net is still obvious, \ie, the percentage gain reaches $3.4\%$ in terms of the Precision score, $3.8\%$ in terms of the Recall score, $4.9\%$ in terms of the F-measure, $29.4\%$ in terms of the MAE score, and $6.7\%$ in terms of the S-measure. It is worth mentioning that our method achieves significant performance improvements when compared with the existing saliency detection methods in optical RSIs. For example, compared with the ASD method \cite{rs4}, the percentage gain reaches $63.9\%$ in terms of F-measure, $90.2\%$ in terms of MAE score, and $61.0\%$ in terms of S-measure.

In Table \ref{tab2_0}, we report the average running time (seconds per image) of different methods on the testing subset of ORSSD dataset and the model size (MB) of deep learning-based methods. The model size is a commonly used metric which indicates the complexity of deep learning-based methods. Since the results of SSD, SPS, and ASD methods are directly provided by authors, we are unable to calculate the running time. As shown, most of deep learning based methods are faster than the traditional methods during the testing phase. As shown in the second row of Table \ref{tab2_0}, the proposed method ranks the second smallest in terms of the model size. All these visual examples and quantitative measures demonstrate the effectiveness and efficiency of the proposed LV-Net.

\begin{table}[!t]
\renewcommand\arraystretch{1.2}
\caption{Comparisons of the Average Running Time (seconds per image) on the Testing Subset of ORSSD Dataset and Model Size (MB).}
\begin{center}
\setlength{\tabcolsep}{0.9mm}{
\begin{tabular}{c|c|c|c|c|c|c|c|c}
\hline
 Methods & DSR & RBD & RRWR & HDCT & DSG & MILPS & RCRR & SSD\\
\hline
 Time & $14.22$ & $0.62$ & $2.91$ & $7.13$ & $1.57$ & $26.34$ & $3.14$ & $-$   \\
\hline
 Model size & $-$ & $-$ & $-$ & $-$ & $-$ & $-$ & $-$ & $-$   \\
\hline\hline
Methods & SPS & ASD & DSS & RADF & R3Net & RFCN & LV-Net & \\
\hline
 Time & $-$ & $-$ & $0.12$ & $0.15$ & $0.48$ & $1.10$ & $0.74$  &\\
\hline
 Model size & $-$ & $-$ & $248$ & $248$ & $142$ & $744$ & $207$  &\\
\hline
\end{tabular}}
\end{center}
\label{tab2_0}
\end{table}
\subsection{Module Analysis}

To demonstrate the improvements obtained by each component in our LV-Net, we conduct the following ablation studies\footnote{The architectures of the compared networks are provided in the supplementary material.}:

\begin{itemize}
  \item \textbf{LV-Net w/o Input-Pyramid}: LV-Net without the input pyramid
  \item \textbf{LV-Net w/o Feature-Pyramid}: LV-Net without the feature pyramid
  \item \textbf{LV-Net w/o L}: LV-Net without the L-shaped module
  \item \textbf{LV-Net w/o Nest}: LV-Net without the nested connections
  \item \textbf{LV-Net w/o Nest+}: LV-Net without the nested connections, but with the skip connections at different levels from encoder pathway to decoder pathway
  \item \textbf{V-Net}: LV-Net without the L-shaped module and the nested connections, but with the skip connections at different levels from encoder pathway to decoder pathway
  \item \textbf{V-Net-D}: V-Net with the double number of feature maps

\end{itemize}

For a fair comparison, we use the same network parameters as the aforementioned settings, except for the \textbf{V-Net-D}. We double the number of feature maps for the V-Net to demonstrate that the good performance of the proposed LV-Net is not on account of the larger hyper-parameters. The quantitative results on the testing subset are reported in Table \ref{tab2}.

In Table \ref{tab2}, the proposed LV-Net achieves the superior performance to other variant networks in terms of all the evaluation metrics, which indicates the advantage of our network architecture. The separate performance of the \textbf{LV-Net w/o Input-Pyramid} and \textbf{LV-Net w/o Feature-Pyramid} is worse than the proposed LV-Net, indicating that the complementary combination of detail features and semantic features is effective and boosts the performance of our network. Moreover, the input pyramid stream can bring more performance gains than the feature pyramid stream because it preserves the original input information, which is significant for the densely connected network structure. Additionally, the performance of \textbf{LV-Net w/o Input-Pyramid} and \textbf{LV-Net w/o Feature-Pyramid} is slightly better than the \textbf{LV-Net w/o L}, which indicates both input pyramid and feature pyramid have positive effects on the LV-Net. The \textbf{LV-Net w/o Nest} and \textbf{LV-Net w/o Nest+} rank the last two, which demonstrates the importance of the nest connections. Besides, the comparisons between the \textbf{LV-Net w/o Nest} and the \textbf{LV-Net w/o Nest+} also indicate that the brute-force skip connections do not introduce obvious performance improvement in the task of salient object detection in optical RSIs. In fact, the final performance is directly related to the learned features. Compared with the \textbf{LV-Net w/o L} and the \textbf{LV-Net w/o Nest}, the LV-Net achieves more competitive performance, indirectly demonstrating that more comprehensive and discriminative feature representations can be learned by the LV-Net than any one of them. The V-Net achieves performance comparable to the \textbf{V-Net-D}, which indicates that enlarging the number of hyper-parameters does not obtain better performance.

\begin{table}[!t]
\renewcommand\arraystretch{1.2}
\caption{Quantitative Evaluation of Ablation Studies on the Testing Subset of ORSSD Dataset.}
\begin{center}
\setlength{\tabcolsep}{3mm}{
\begin{tabular}{c|c|c|c}
\hline
  & $F_{\beta}$ & MAE & $S_m$\\
\hline\hline
LV-Net w/o Input-Pyramid & $0.8297$ & $0.0231$ & $0.8713$\\
\hline
LV-Net w/o Feature-Pyramid & $0.8335$ & $0.0227$ & $0.8784$\\
\hline
LV-Net w/o L & $0.8248$ & $0.0240$ & $0.8684$ \\
\hline
LV-Net w/o Nest& $0.7672$ & $0.0280$ & $0.8299$ \\
\hline
LV-Net w/o Nest+& $0.7821$ & $0.0334$ & $0.8288$ \\
\hline
V-Net & $0.8090$ & $0.0297$ & $0.8446$ \\
\hline
V-Net-D & $0.8089$ & $0.0302$ & $0.8429$  \\
\hline
LV-Net & \textbf{0.8414} & \textbf{0.0207} & \textbf{0.8815} \\
\hline
\end{tabular}}
\end{center}
\label{tab2}
\end{table}

\begin{figure}[!t]
\centering
\centerline{\includegraphics[width=1\linewidth]{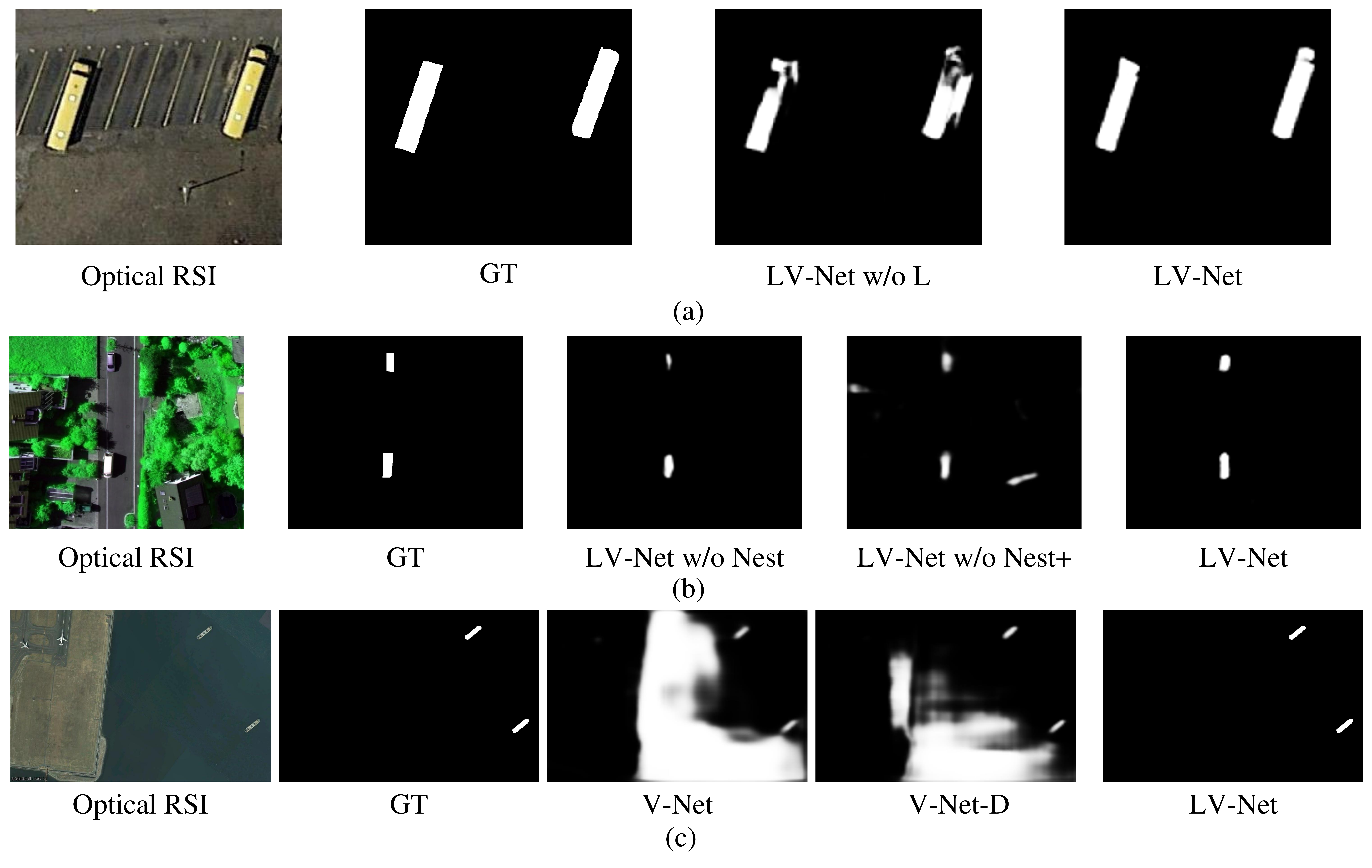}}
\caption{Visual examples of module analysis. (a) Evaluation of L-shaped module. (b) Evaluation of nested connections. (c) Evaluation of V-Net module.}
\label{fig8}
\end{figure}

In addition, some visual comparison results are shown in Fig. \ref{fig8}. Observing Fig. \ref{fig8}(a), we can see that the detected salient objects by the proposed LV-Net are more clear and sharper than the \textbf{LV-Net w/o L} variation, such as the boundary and edge. In Fig. \ref{fig8}(b), the salient objects by the \textbf{LV-Net w/o Nest} variation have no clear and complete boundary due to the lack of the low-level detail features. In contrast, the result of \textbf{LV-Net w/o Nest+} obtains more complete boundary; however, it introduces the background noise through the brute-force skip connections. As can be seen from Fig. \ref{fig8}(c), the backbone of the V-Net fails to filter out the backgrounds effectively when compared with the proposed LV-Net, which further indicates that the importance of the L-shaped module and the nested connections used in the proposed LV-Net.

In summary, the ablation studies demonstrate that a) the competitive performance of the proposed LV-Net benefits from the introduction of both the two-stream pyramid and the nested connections; b) the  backbone of the V-Net is not suitable for salient object detection in optical RSIs with complicated backgrounds because of the use of brute-force skip connections; and c) reasonable network design can achieve better performance than only enlarging the hyper-parameters.

\begin{table}[!t]
\renewcommand\arraystretch{1.2}
\caption{Quantitative Evaluation of the Effects of Network Parameter Settings on the Testing Subset of ORSSD Dataset.}
\begin{center}
\setlength{\tabcolsep}{3mm}{
\begin{tabular}{c|c|c|c}
\hline
  & $F_{\beta}$ & MAE & $S_m$\\
\hline\hline
LV-Net-8-16& $0.8080$ & $0.0251$ & $0.8532$ \\
\hline
LV-Net-16-32 & $0.8327$ & $0.0213$ & $0.8782$ \\
\hline
LV-Net-3Scales& $0.7963$ & $0.0348$ & $0.8168$ \\
\hline
LV-Net-4Scales& $0.8385$ & $0.0249$ & $ 0.8634$ \\
\hline
LV-Net-S-CU & $0.8379$ & $0.0212$ & $0.8752$ \\
\hline
LV-Net& $\textbf{0.8414}$ & $\textbf{0.0207}$ & $\textbf{0.8815}$ \\
\hline
\end{tabular}}
\end{center}
\label{tab4}
\end{table}

\subsection{Parameter Analysis}
We conduct experiments to analyze the effects of network parameter settings, including the numbers of filters and scales, and the kernel size. The quantitative comparison results on the testing subset are reported in Table \ref{tab4}.

First, based on the default settings of the number of output features (\ie, $32\times2^k$ for each convolutional layer in the M-CU, and $64\times2^i$ for each convolutional layer in the encoder-decoder module), we reduce the number of output features to ($16\times2^k$, $32\times2^i$) and ($8\times2^k$, $16\times2^i$) for comparisons and denote them as \textbf{LV-Net-16-32} and \textbf{LV-Net-8-16}, respectively. Here, $k$ and $i$ index the down-sampling layer along the input image and encoder pathway, respectively. Second, we analyze how the number of scales affects the performance. To be specific, we remove the branch of the smallest scale (\ie, reducing $5$ scales to $4$ scales). Then, we adjust the output of the CU$_{(0,3)}$ as the final saliency map and denote it as \textbf{LV-Net-4Scales}. Similarly, we reduce $5$ scales of the proposed LV-Net to $3$ scales (\ie, the output of CU$_{(0,2)}$ as the final saliency map) and denote it as \textbf{LV-Net-3Scales}. Third, we change the multi-scale convolution operations in the M-CU to single-scale convolution operation (\ie, $3\times3$, $3\times3$, and $3\times3$) and denote it as \textbf{LV-Net-S-CU}.

As reported in Table \ref{tab4}, as the decrease of the number of output features, the values of the F-measure and S-measure decrease (the values of the MAE increase). Such results comply with the common conclusion that more features can improve the performance of deep networks to some extent. In addition, we can see that as the number of scales decreases, the performance of the network decreases. The reason is that more scales will contain more useful information, which is beneficial to the learning of more comprehensive feature representations for saliency detection. It also indicates that the multi-scale input and more nested connections can obtain gains. From the results of \textbf{LV-Net-S-CU}, we can see that the multi-scale convolutional layers in the M-CU are helpful for the improvement of the proposed LV-Net.

\section{Conclusion}\label{sec6}
In this paper, we proposed the LV-Net for salient object detection in optical RSIs. Benefiting from both the two-stream pyramid module and the nested connections, the proposed LV-Net can accurately locate the salient objects with diverse scales and effectively suppress the cluttered backgrounds. Moreover, we constructed an optical RSI dataset for salient object detection with pixel-wise annotation. Experiments demonstrate the proposed method significantly outperforms the state-of-the-art methods both qualitatively and quantitatively. The module analysis and parameter discussion demonstrate the effectiveness of each designed component and the parameter settings in the proposed LV-Net.

To further improve the edge sharpness and spatial consistency of the salient regions, we will add a context module at the end of the proposed LV-Net to capture more contextual information. We will also extend the ORSSD dataset and periodically update the results for noticeable salient object detection methods on it. Moreover, some extension works of saliency detection, such as co-saliency detection \cite{cosal1,cosal2,cosal3,cosal4}, will be further extended to optical RSIs in the future.

\par

\ifCLASSOPTIONcaptionsoff
  \newpage
\fi

{
\bibliographystyle{IEEEtran}
\bibliography{ref}

\begin{thebibliography}{10}
\providecommand{\url}[1]{#1}
\csname url@samestyle\endcsname
\providecommand{\newblock}{\relax}
\providecommand{\bibinfo}[2]{#2}
\providecommand{\BIBentrySTDinterwordspacing}{\spaceskip=0pt\relax}
\providecommand{\BIBentryALTinterwordstretchfactor}{4}
\providecommand{\BIBentryALTinterwordspacing}{\spaceskip=\fontdimen2\font plus
\BIBentryALTinterwordstretchfactor\fontdimen3\font minus
  \fontdimen4\font\relax}
\providecommand{\BIBforeignlanguage}[2]{{%
\expandafter\ifx\csname l@#1\endcsname\relax
\typeout{** WARNING: IEEEtran.bst: No hyphenation pattern has been}%
\typeout{** loaded for the language `#1'. Using the pattern for}%
\typeout{** the default language instead.}%
\else
\language=\csname l@#1\endcsname
\fi
#2}}
\providecommand{\BIBdecl}{\relax}
\BIBdecl

\bibitem{R1}
W.~Wang, J.~Shen, R.~Yang, and F.~Porikli, ``Saliency-aware video object
  segmentation,'' \emph{IEEE Trans. Patt. Anal. Mach. Intell.}, vol.~40, no.~1,
  pp. 20--33, Jan. 2018.

\bibitem{R2}
Y.~Fang, Z.~Chen, W.~Lin, and C.-W. Lin, ``Saliency detection in the compressed
  domain for adaptive image retargeting,'' \emph{IEEE Trans. Image Process.},
  vol.~21, no.~9, pp. 3888--3901, Sep. 2012.

\bibitem{R3}
X.~Cao, C.~Zhang, H.~Fu, X.~Guo, and Q.~Tian, ``Saliency-aware nonparametric
  foreground annotation based on weakly labeled data,'' \emph{IEEE Trans.
  Neural Netw. Learn. Syst.}, vol.~27, no.~6, pp. 1253--1265, Jun. 2016.

\bibitem{R4}
W.~Wang, J.~Shen, Y.~Yu, and K.-L. Ma, ``Stereoscopic thumbnail creation via
  efficient stereo saliency detection,'' \emph{IEEE Trans. Vis. Comput. Graph},
  vol.~23, no.~8, pp. 2014--2027, Aug. 2017.

\bibitem{R5}
K.~Gu, S.~Wang, H.~Yang, W.~Lin, G.~Zhai, X.~Yang, and W.~Zhang,
  ``Saliency-guided quality assessment of screen content images,'' \emph{IEEE
  Trans. Multimedia}, vol.~18, no.~6, pp. 1098--1110, Jun. 2016.

\bibitem{R6}
H.~Jacob, F.~Padua, A.~Lacerda, and A.~Pereira, ``Video summarization approach
  based on the emulation of bottom-up mechanisms of visual attention,''
  \emph{J. Intell. Information Syst.}, vol.~49, no.~2, pp. 193--211, Feb. 2017.

\bibitem{RERVIEW}
R.~Cong, J.~Lei, H.~Fu, M.-M. Cheng, W.~Lin, and Q.~Huang, ``Review of visual
  saliency detectioin with comprehensive information,'' \emph{IEEE Trans.
  Circuits Syst. Video Technol}, vol.~PP, no.~99, pp. 1--19, 2018.

\bibitem{add1}
H.~Lin, Z.~Shi, and Z.~Zou, ``Fully convolutional network with task
  partitioning for inshore ship detection in optical remote sensing images,''
  \emph{IEEE Geosci. Remote Sens. Lett.}, vol.~14, no.~10, pp. 1665--1669, Oct.
  2017.

\bibitem{rs4_add}
C.~Dong, J.~Liu, and F.~Xu, ``Ship detection in optical remote sensing images
  based on saliency and a rotation-invariant descriptor,'' \emph{Remote Sens.},
  vol.~10, no.~3, pp. 1--19, Mar. 2018.

\bibitem{add2}
Z.~Xiao, Y.~Gong, Y.~Long, D.~Li, X.~Wang, and H.~Liu, ``Airport detection
  based on a multiscale fusion feature for optical remote sensing images,''
  \emph{IEEE Geosci. Remote Sens. Lett.}, vol.~14, no.~9, pp. 1469--1473, Sep.
  2017.

\bibitem{rs0}
J.~Han, D.~Zhang, G.~Cheng, L.~Guo, and J.~Ren, ``Object detection in optical
  remote sensing images based on weakly supervised learning and high-level
  feature learning,'' \emph{IEEE Trans. Geosci. Remote Sens.}, vol.~53, no.~6,
  pp. 3325--3337, Jun. 2015.

\bibitem{add3}
G.~Cheng and J.~Han, ``A survey on object detection in optical remote sensing
  images,'' \emph{ISPRS Journal of Photogrammetry and Remote Sensing}, vol.
  117, pp. 11--28, Jul. 2016.

\bibitem{add4}
G.~Cheng, J.~Han, P.~Zhou, and L.~Guo, ``Multi-class geospatial object
  detection and geographic image classification based on collection of part
  detectors,'' \emph{ISPRS Journal of Photogrammetry and Remote Sensing},
  vol.~98, pp. 119--132, Dec. 2014.

\bibitem{add5}
K.~Li, G.~Cheng, S.~Bu, and X.~You, ``Rotation-insensitive and
  context-augmented object detection in remote sensing images,'' \emph{IEEE
  Trans. Geosci. Remote Sens.}, vol.~56, no.~4, pp. 2337--2348, Apr. 2018.

\bibitem{RCRR}
Y.~Yuan, C.~Li, J.~Kim, W.~Cai, and D.~D. Feng, ``Reversion correction and
  regularized random walk ranking for saliency detection,'' \emph{IEEE Trans.
  Image Process.}, vol.~27, no.~3, pp. 1311--1322, Mar. 2018.

\bibitem{R3Net}
Z.~Deng, X.~Hu, L.~Zhu, X.~Xu, J.~Qin, G.~Han, and P.-A. Heng, ``R$^{3}${N}et:
  Recurrent residual refinement network for saliency detection,'' in
  \emph{Proc. IJCAI}, Jul. 2018, pp. 684--690.

\bibitem{RC}
M.-M. Cheng, G.-X. Zhang, N.~J. Mitra, X.~Huang, and S.-M. Hu, ``Global
  contrast based salient region detection,'' in \emph{Proc. CVPR}, Jun. 2011,
  pp. 409--416.

\bibitem{RBD}
W.~Zhu, S.~Liang, Y.~Wei, and J.~Sun, ``Saliency optimization from robust
  background detection,'' in \emph{Proc. CVPR}, Jun. 2014, pp. 2814--2821.

\bibitem{DCLC}
L.~Zhou, Z.~Yang, Q.~Yuan, Z.~Zhou, and D.~Hu, ``Salient region detection via
  integrating diffusion-based compactness and local contrast,'' \emph{IEEE
  Trans. Image Process.}, vol.~24, no.~11, pp. 3308--3320, Nov. 2015.

\bibitem{DSR}
X.~Li, H.~Lu, L.~Zhang, X.~Ruan, and M.-H. Yang, ``Saliency detection via dense
  and sparse reconstruction,'' in \emph{Proc. ICCV}, Oct. 2013, pp. 2976--2983.

\bibitem{SMD}
H.~Peng, B.~Li, H.~Ling, W.~Hua, W.~Xiong, and S.~Maybank, ``Salient object
  detection via structured matrix decomposition,'' \emph{IEEE Trans. Patt.
  Anal. Mach. Intell.}, vol.~39, no.~4, pp. 818--832, Apr. 2017.

\bibitem{SuperCNN}
S.~He, R.~W. Lau, W.~Liu, Z.~Huang, and Q.~Yang, ``{SuperCNN}: A superpixelwise
  convolutional neural network for salient object detection,'' \emph{Int. J.
  Comput. Vis.}, vol. 115, no.~3, pp. 330--344, Mar. 2015.

\bibitem{DCL}
G.~Li and Y.~Yu, ``Deep contrast learning for salient object detection,'' in
  \emph{Proc. CVPR}, Jun. 2016, pp. 478--487.

\bibitem{DSS}
Q.~Hou, M.-M. Cheng, X.~Hu, A.~Borji, Z.~Tu, and P.~Torr, ``Deeply supervised
  salient object detection with short connections,'' in \emph{Proc. CVPR}, Jun.
  2017, pp. 5300--5309.

\bibitem{RADF}
X.~Hu, L.~Zhu, J.~Qin, C.-W. Fu, and P.-A. Heng, ``Recurrently aggregating deep
  features for salient object detection,'' in \emph{Proc. AAAI}, Feb. 2018, pp.
  6943--6950.

\bibitem{UCF}
P.~Zhang, D.~Wang, H.~Lu, H.~Wang, and B.~Yin, ``Learning uncertain
  convolutional features for accurate saliency detection,'' in \emph{Proc.
  ICCV}, Oct. 2017, pp. 212--221.

\bibitem{DSCLRCN}
N.~Liu and J.~Han, ``A deep spatial contextual long-term recurrent
  convolutional network for saliency detection,'' \emph{IEEE Trans. Image
  Process.}, vol.~27, no.~7, pp. 3264--3274, Jul. 2018.

\bibitem{RFCN}
L.~Wang, L.~Wang, H.~Lu, P.~Zhang, and X.~Ruan, ``Salient object detection with
  recurrent fully convolutional networks,'' \emph{IEEE Trans. Patt. Anal. Mach.
  Intell.}, vol.~PP, no.~99, pp. 1--13, 2018.

\bibitem{rs5}
D.~Zhao, J.~Wang, J.~Shi, and Z.~Jiang, ``Sparsity-guided saliency detection
  for remote sensing images,'' \emph{J. Applied Remote Sens.}, vol.~9, pp.
  1--14, Sep. 2015.

\bibitem{rs1}
E.~Li, S.~Xu, W.~Meng, and X.~Zhang, ``Building extraction from remotely sensed
  images by integrating saliency cue,'' \emph{IEEE J. Sel. Topics Appl. Earth
  Observ.}, vol.~10, no.~3, pp. 906--919, Mar. 2017.

\bibitem{rs2}
L.~Ma, B.~Du, H.~Chen, and N.~Q. Soomro, ``Region-of-interest detection via
  superpixel-to-pixel saliency analysis for remote sensing image,'' \emph{IEEE
  Geosci. Remote Sens. Lett.}, vol.~13, no.~12, pp. 1752--1756, Dec. 2016.

\bibitem{rs3}
T.~Li, J.~Zhang, X.~Lu, and Y.~Zhang, ``{SDBD}: A hierarchical
  region-of-interest detection approach in large-scale remote sensing image,''
  \emph{IEEE Geosci. Remote Sens. Lett.}, vol.~14, no.~5, pp. 699--703, May
  2017.

\bibitem{rs4}
Q.~Zhang, L.~Zhang, W.~Shi, and Y.~Liu, ``Airport extraction via complementary
  saliency analysis and saliency-oriented active contour model,'' \emph{IEEE
  Geosci. Remote Sens. Lett.}, vol.~15, no.~7, pp. 1085--1089, Jul. 2018.

\bibitem{ReLU}
A.~Krizhevsky, I.~Sutskever, and G.~Hinton, ``Imagenet classification with deep
  convolutional neural networks,'' in \emph{Proc. NIPS}, Dec. 2012, pp.
  1097--1105.

\bibitem{DenseNet}
G.~Huang, Z.~Liu, L.~V.~D. Maaten, and K.~Weinberger, ``Densely connected
  convolutional networks,'' in \emph{Proc. CVPR}, Jun. 2017, pp. 4548--4557.

\bibitem{U-Net}
O.~Ronnerberger, P.~Fischer, and T.~Brox, ``U-net: Convolutional networks for
  biomedical image segmentation,'' in \emph{Proc. MICCAI}, Oct. 2015, pp.
  234--241.

\bibitem{ELD-Net}
G.~Lee, Y.~Tai, and J.~Kim, ``Eld-net: An efficient deep learning architecture
  for accurate saliency detection,'' \emph{IEEE Trans. Patt. Anal. Mach.
  Intell.}, vol.~40, no.~7, pp. 1599--1610, Jul. 2018.

\bibitem{rs0_1}
G.~Cheng, P.~Zhou, and J.~Han, ``Learning rotation-invariant convolutional
  neural networks for object detection in {VHR} optical remote sensing
  images,'' \emph{IEEE Trans. Geosci. Remote Sens.}, vol.~54, no.~12, pp.
  7405--7415, Dec. 2016.

\bibitem{AID}
G.-S. Xia, J.~Hu, F.~Hu, B.~Shi, X.~Bai, Y.~Zhong, L.~Zhang, and X.~Lu, ``Aid:
  A benchmark dataset for performance evaluation of aerial scene
  classification,'' \emph{IEEE Trans. Geosci. Remote Sens.}, vol.~55, no.~7,
  pp. 3965--3981, Jul. 2017.

\bibitem{LEVIR}
Z.~Zou and Z.~Shi, ``Random access memories: A new paradigm for target
  detection in high resolution aerial remote sensing images,'' \emph{IEEE
  Trans. Image Process.}, vol.~27, no.~3, pp. 1100--1111, Mar. 2018.

\bibitem{rs0_3}
G.~Cheng, J.~Han, and X.~Lu, ``Remote sensing image scene classification:
  Benchmark and state of the art,'' \emph{Proc. IEEE}, vol. 105, no.~10, pp.
  1865--1883, Oct. 2017.

\bibitem{Fmeasure2}
R.~Achanta, S.~Hemami, F.~Estrada, and S.~Ssstrunk, ``Frequency-tuned salient
  region detection,'' in \emph{Proc. CVPR}, Jun. 2009, pp. 1597--1604.

\bibitem{Fmeasure}
A.~Borji, M.-M. Cheng, H.~Jiang, and J.~Li, ``Salient object detection: A
  benchmark,'' \emph{IEEE Trans. Image Process.}, vol.~24, no.~12, pp.
  5706--5722, Dec. 2015.

\bibitem{MAE}
D.~Zhang, H.~Fu, J.~Han, A.~Borji, and X.~Li, ``A review of co-saliency
  detection algorithms: Fundamentals, applications, and challenges,'' \emph{ACM
  Trans. on Intell. Syst. and Technol.}, vol.~9, no.~4, pp. 1--31, Apr. 2018.

\bibitem{S-measure}
D.-P. Fan, M.-M. Cheng, Y.~Liu, T.~Li, and A.~Borji, ``Structure-measure: A new
  way to evaluate foreground maps,'' in \emph{Proc. ICCV}, Oct. 2017, pp.
  4548--4557.

\bibitem{Xavier}
X.~Glorot and Y.~Bengio, ``Understanding the difficulty of training deep
  feedforward neural networks,'' in \emph{Proc. AISTATS}, May 2010, pp.
  249--256.

\bibitem{Kingma2014}
D.~Kingma and J.~Ba, ``Adam: A method for stochastic optimization,''
  \emph{arXiv preprint arXiv:1412.6980}, Jan. 2017.

\bibitem{RRWR}
C.~Li, Y.~Yuan, W.~Cai, Y.~Xia, and D.~Feng, ``Robust saliency detection via
  regularized random walks ranking,'' in \emph{Proc. CVPR}, Jun. 2015, pp.
  2710--2717.

\bibitem{HDCT}
J.~Kim, D.~Han, Y.-W. Tai, and J.~Kim, ``Salient region detection via
  high-dimensional color transform and local spatial support,'' \emph{IEEE
  Trans. Image Process.}, vol.~25, no.~1, pp. 9--23, Jan. 2015.

\bibitem{DSG}
L.~Zhou, Z.~Yang, Z.~Zhou, and D.~Hu, ``Salient region detection using
  diffusion process on a two-layer sparse graph,'' \emph{IEEE Trans. Image
  Process.}, vol.~26, no.~12, pp. 5882--5894, Dec. 2017.

\bibitem{MILPS}
F.~Huang, J.~Qi, H.~Lu, L.~Zhang, and X.~Ruan, ``Salient object detection via
  multiple instance learning,'' \emph{IEEE Trans. Image Process.}, vol.~26,
  no.~4, pp. 1911--1922, Apr. 2017.

\bibitem{cosal1}
J.~Han, G.~Cheng, Z.~Li, and D.~Zhang, ``A unified metric learning-based for
  co-saliency detection framework,'' \emph{IEEE Trans. Circuits Syst. Video
  Technol.}, vol.~28, no.~10, pp. 2473--2483, Oct. 2018.

\bibitem{cosal2}
R.~Cong, J.~Lei, H.~Fu, Q.~Huang, X.~Cao, and C.~Hou, ``Co-saliency detection
  for {RGBD} images based on multi-constraint feature matching and cross label
  propagation,'' \emph{IEEE Trans. Image Process.}, vol.~27, no.~2, pp.
  568--579, Feb. 2018.

\bibitem{cosal3}
R.~Cong, J.~Lei, H.~Fu, W.~Lin, Q.~Huang, X.~Cao, and C.~Hou, ``An iterative
  co-saliency framework for {RGBD} images,'' \emph{IEEE Trans. on Cybern.},
  vol.~49, no.~1, pp. 233--246, Jan. 2019.

\bibitem{cosal4}
J.~Han, D.~Zhang, G.~Cheng, N.~Liu, and D.~Xu, ``Advanced deep-learning
  techniques for salient and category-specific object detection: A survey,''
  \emph{IEEE Signal Process. Mag.}, vol.~35, no.~1, pp. 84--100, Jan. 2018.

\end{thebibliography}
}

\begin{IEEEbiography}[{\includegraphics[width=1in,height=1.25in,clip,keepaspectratio]{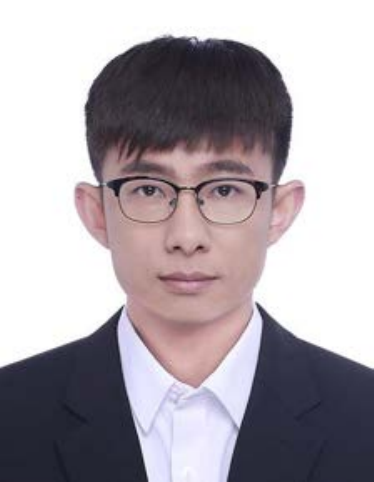}}]{Chongyi Li}
received his Ph.D. degree with the School of Electrical and Information Engineering, Tianjin University, Tianjin, China in June 2018. Now, he is a Postdoc Research Fellow at the Department of Computer Science, City University of Hong Kong (CityU), Hong Kong. His current research focuses on image processing, computer vision, and deep learning, particularly in the domains of image restoration and enhancement.
\end{IEEEbiography}
\vspace{-10 mm}
\begin{IEEEbiography}[{\includegraphics[width=1in,height=1.25in,clip,keepaspectratio]{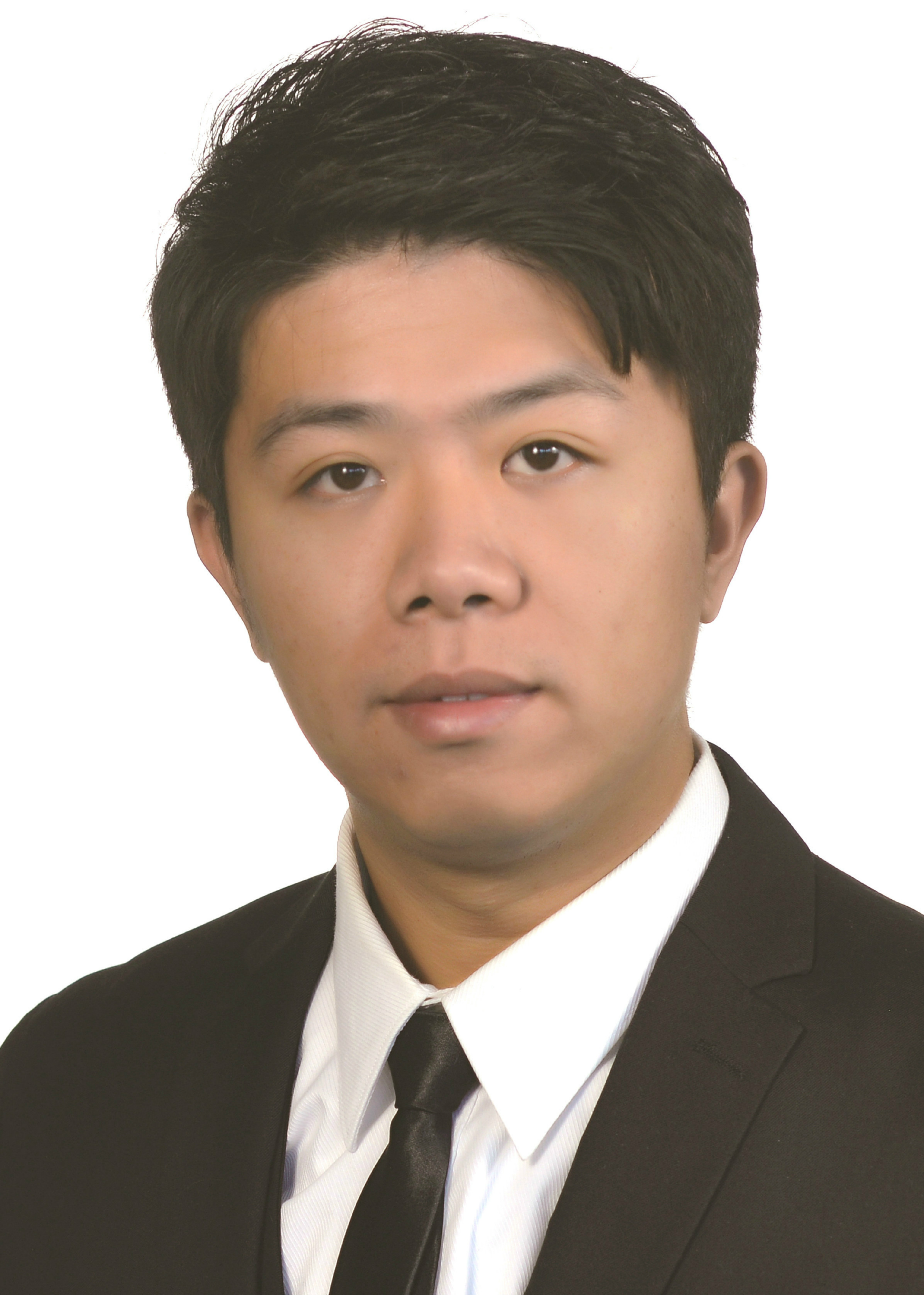}}]{Runmin Cong}
received the Ph.D. degree in information and communication engineering from Tianjin University, Tianjin, China, in 2019.\par
He is currently an Associate Professor with the Institute of Information Science, Beijing Jiaotong University, Beijing, China. He was a research student/staff at Nanyang Technological University (NTU), Singapore, and City University of Hong Kong (CityU), Hong Kong. He won the Best Student Paper Runner-Up at IEEE ICME in 2018. His research interests include computer vision, image processing, saliency detection, and 3-D imaging.
\end{IEEEbiography}
\vspace{-10 mm}
\begin{IEEEbiography}[{\includegraphics[width=1in,height=1.25in,clip,keepaspectratio]{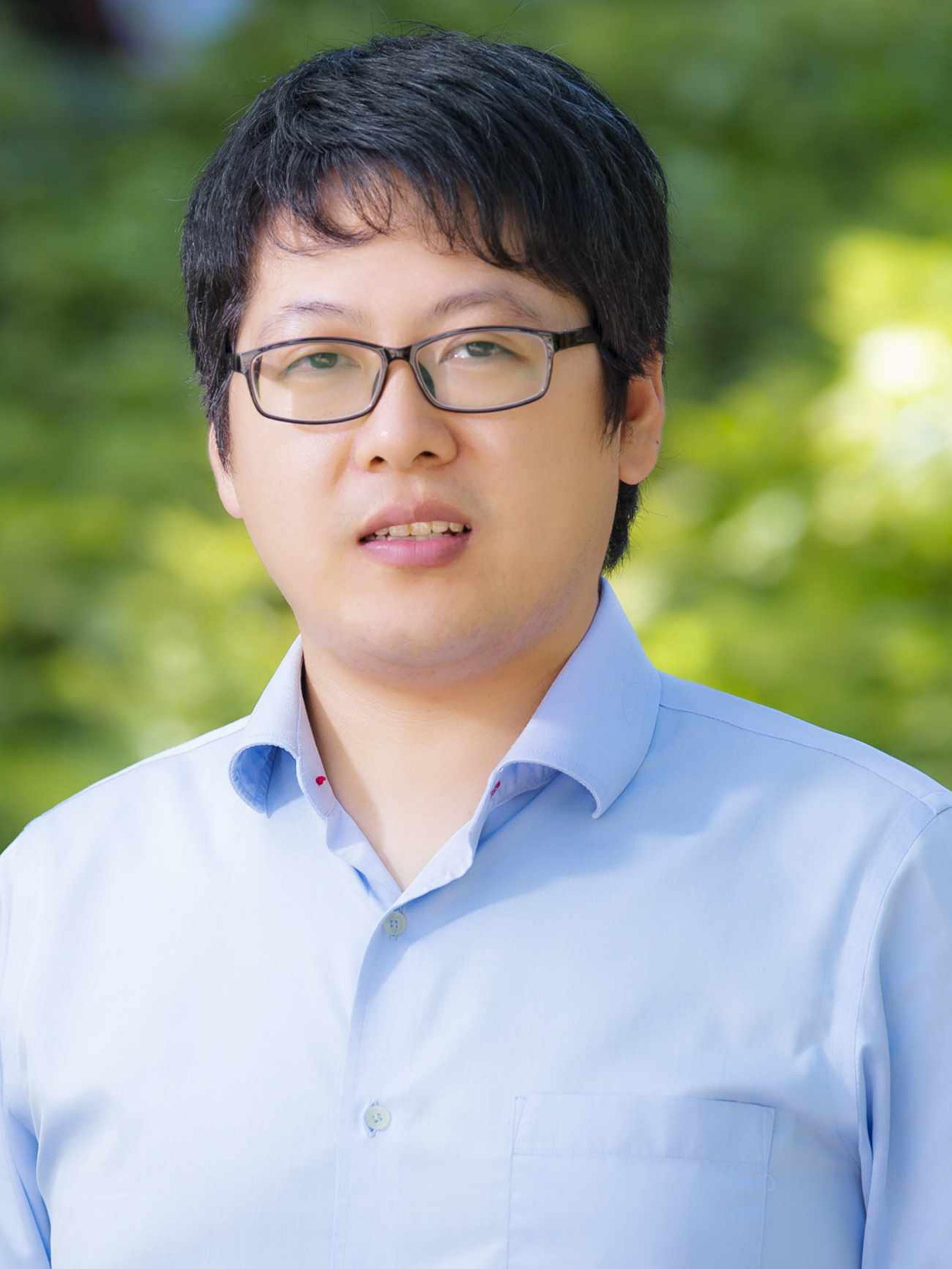}}]{Junhui Hou}
(S'13-M'16) received the B.Eng. degree in information engineering (Talented Students Program) from the South China University of Technology, Guangzhou, China, in 2009, the M.Eng. degree in signal and information processing from Northwestern Polytechnical University, Xian, China, in 2012, and the Ph.D. degree from the School of Electrical and Electronic Engineering, Nanyang Technological University, Singapore, in 2016.\par
He has been an Assistant Professor with the Department of Computer Science, City University of Hong Kong, since 2017. His current research interests include multimedia signal processing, such as adaptive image/video representations and analysis (RGB/depth/light field/hyperspectral), static/dynamic 3D geometry representations and processing (mesh/point cloud/MoCap), and discriminative modeling for clustering/classification.\par
Dr. Hou was a recipient of the Prestigious Award from the Chinese Government for Outstanding Self-Financed Students Abroad, China Scholarship Council, in 2015, and the Early Career Award from the Hong Kong Research Grants Council in 2018. He currently serves as an Associate Editor for \emph{The Visual Computer} and an Area Editor for \emph{Signal Processing:Image Communication}.
\end{IEEEbiography}
\vspace{-10 mm}
\begin{IEEEbiography}[{\includegraphics[width=1in,height=1.25in,clip,keepaspectratio]{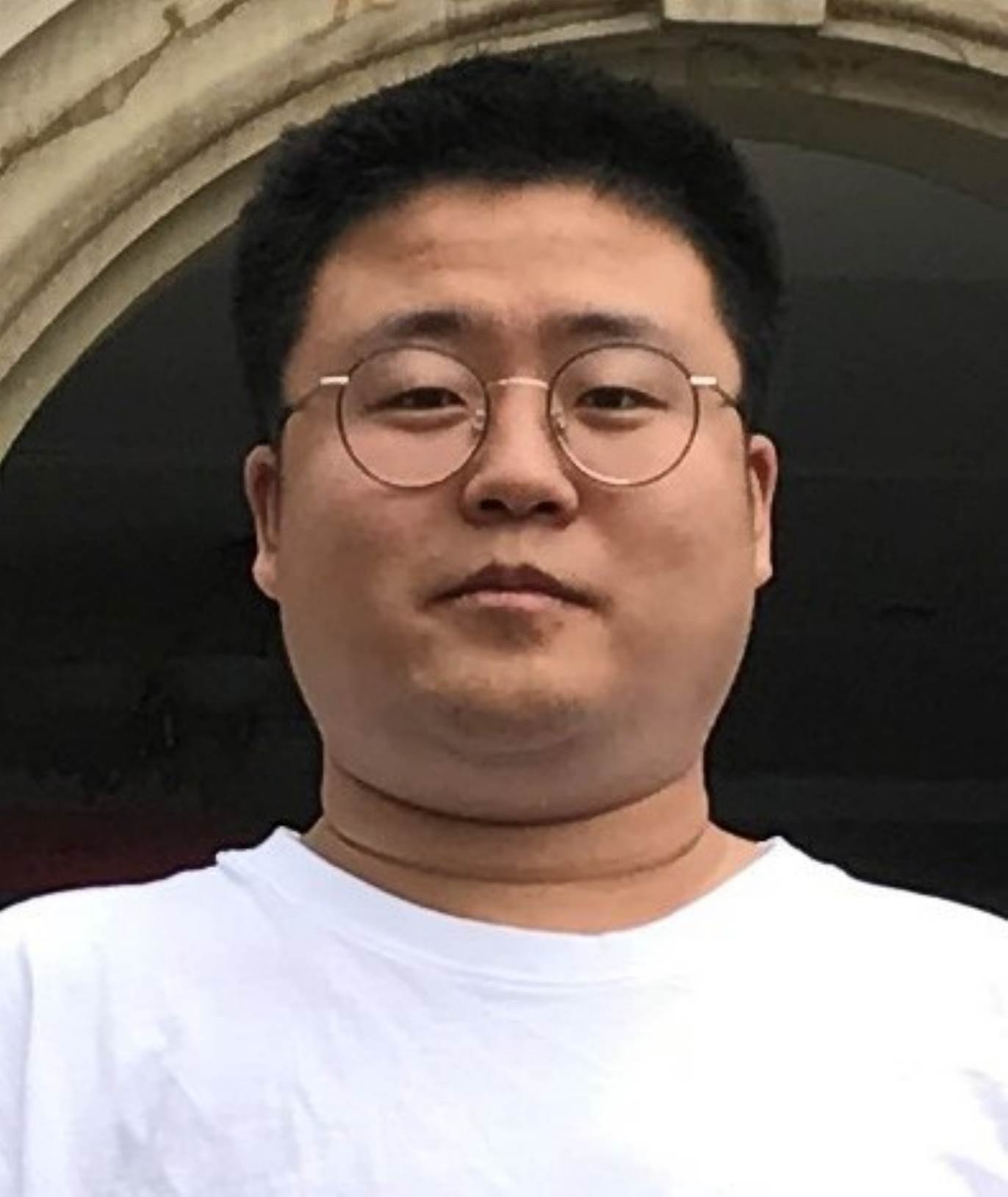}}]{Sanyi Zhang}
is a Ph.D. candidate at School of Electrical and Information Engineering, Tianjin University, China. He received his B.E. and M.E. degrees in computer science from Taiyuan University of Technology, China. His current research interests include computer vision and clothing attribute learning.
\end{IEEEbiography}
\vspace{-10 mm}
\begin{IEEEbiography}[{\includegraphics[width=1in,height=1.25in,clip,keepaspectratio]{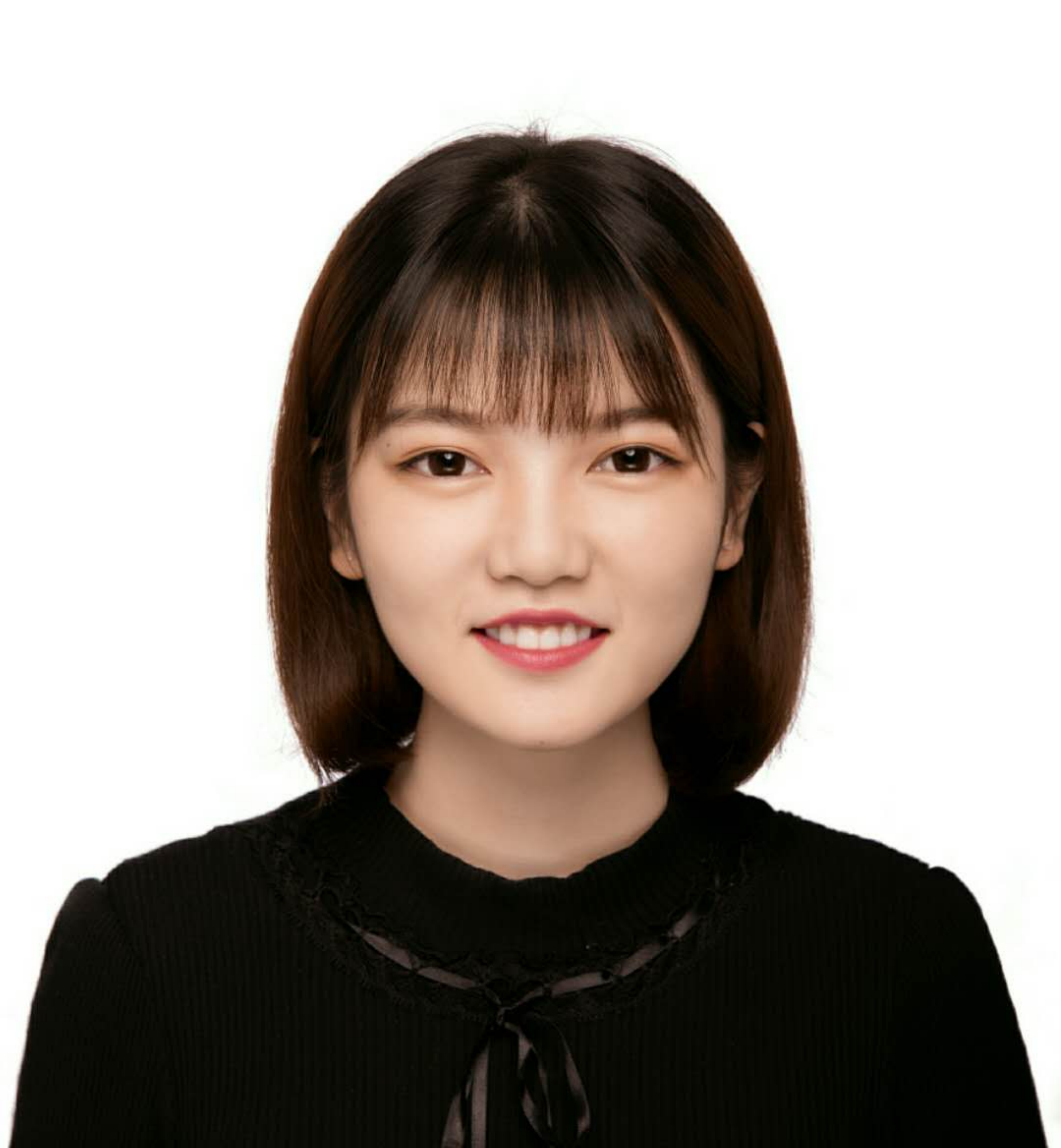}}]{Yue Qian}
received the B.S and M.Phil degree in Mathematics from The Chinese Univerisity of Hong Kong in 2014 and 2016 respectively. Currently, she is a Ph.D student at City Unversity of Hong Kong studying Computer Science. Her research interests include image processing and computer vision.
\end{IEEEbiography}
\vspace{-10 mm}
\begin{IEEEbiography}[{\includegraphics[width=1in,height=1.25in,clip,keepaspectratio]{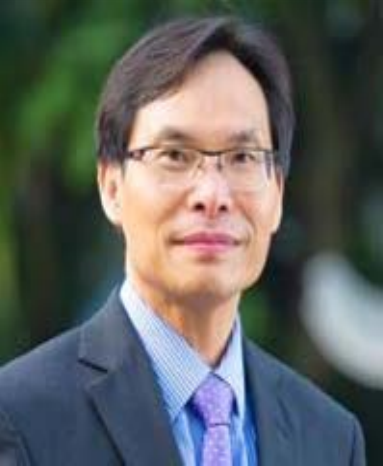}}]{Sam Kwong}
(M'93-SM'04-F'13) received the B.S. degree in electrical engineering from The State University of New York at Buffalo in 1983, the M.S. degree from the University of Waterloo, Waterloo, ON, Canada, in 1985, and the Ph.D. degree from the University of Hagen, Germany, in 1996. From 1985 to 1987, he was a Diagnostic Engineer with Control Data Canada. He joined Bell Northern Research Canada as a member of Scientific Staff. In 1990,
he became a Lecturer with the Department of Electronic Engineering, City University of Hong Kong, where he is currently a Chair Professor with the Department of Computer Science. His research interests are video and image coding and evolutionary algorithms.
\end{IEEEbiography}

\end{document}